%% file: paper.tex
\documentclass[runningheads]{llncs}
\usepackage{graphicx}
\usepackage{svg}
\usepackage{amsfonts}
\usepackage{amsmath}
\usepackage{calrsfs}
\usepackage[ruled,linesnumbered,noend]{algorithm2e}
\usepackage[noabbrev,capitalise]{cleveref}
\usepackage{subfig}
\usepackage[framemethod=tikz]{mdframed}

\usepackage{xcolor}

\DeclareMathAlphabet{\pazocal}{OMS}{zplm}{m}{n}
\newcommand{\Lb}{\pazocal{L}}
\newcommand{\LabellingFunction}{$\Lb: S \rightarrow 2^{P}$}
\newcommand{\LTLf}{LTL\textsubscript{\textit{f}}}
\newcommand{\LDLf}{LDL\textsubscript{\textit{f}}}
\newcommand{\DFA}{$A_{\varphi} = \langle 2^{P}, Q, \delta, F, q_{0} \rangle$}
\newcommand{\RDPTransitionQuadruples}{$\{(\varphi_{i}, a, P_{i}^{\prime}, \pi_{i}(P^{\prime}))\;|\;i \in I_{a}\}$}
\newcommand{\RewardFunction}{$R: S \times A \times S \rightarrow \mathbb{R}$}
\newcommand{\RewardShapingFunction}{$R^{\prime}(s, a, s^{\prime}) + F(s, a, s^{\prime}) \rightarrow \mathbb{R}$}
\newcommand{\ShapingRewardFunction}{$F(s, a, s^{\prime})$}
\newcommand{\PotentialRewardFunction}{$F(s, a, s^{\prime}) \rightarrow \gamma\Phi(s^{\prime}) - \Phi(s)$ for some $\Phi : S \rightarrow \mathbb{R}$}

\newcounter{experiment}[part]
\newenvironment{experiment}[3][]{
    \refstepcounter{experiment}\label{experiment:#2}
    \subsection{Experiment~\theexperiment: #3}
}{}
\crefname{experiment}{Experiment}{Experiments}

\begin{document}
\title{Regular Decision Processes for Grid Worlds}
%
%
\author{Nicky Lenaers \and Martijn van Otterlo}
\authorrunning{N. Lenaers \and M. van Otterlo}
%
\institute{Open University, The Netherlands}
\maketitle              
\begin{abstract}
Markov decision processes are typically used for sequential decision making under uncertainty. For many aspects however, ranging from \emph{constrained} or \emph{safe} specifications to various kinds of temporal (\emph{non-Markovian}) dependencies in task and reward structures, extensions are needed. To that end, in recent years interest has grown into combinations of reinforcement learning and temporal logic, that is, combinations of flexible behavior learning methods with robust verification and guarantees. In this paper we describe an experimental investigation of the recently introduced \emph{regular decision processes} that support both non-Markovian reward functions as well as transition functions. In particular, we provide a tool chain for regular decision processes, algorithmic extensions relating to online, incremental learning, an empirical evaluation of model-free and model-based solution algorithms, and applications in regular, but non-Markovian, grid worlds.

\noindent\keywords{sequential decisions \and safe reinforcement learning \and non-Markovian dynamics \and regular decision process \and linear temporal logic}
\end{abstract}

%
%
\section{Introduction}
\emph{Sequential decision making under uncertainty}, often simply denoted by its core algorithmic subfield \emph{reinforcement learning} (RL) \cite{van2009logic,wiering2012reinforcement,10.5555/3312046}, has been showing a huge amount of progress the last decades. Among the recent breakthroughs is the progression of DeepMind's RL methods solving the board game Go \cite{silver2016mastering}, chess, Atari computer games, the real-time strategy game StarCraft II, and lately chip design \cite{mirhoseini2021graph}. The algorithms employ combinations of (Monte Carlo) planning and value function approximation using deep neural networks.

Underlying typical RL systems is the \emph{Markov decision process} (MDP) \cite{10.1002/9780470316887} in which \emph{states} carry all necessary information to choose (optimal) \emph{actions}. The \emph{Markov property} dictates that given the present, the future is \emph{independent} of the past. To scale to more complex problems, one can exploit \emph{structure} in the space of state(-action) spaces, or policies or value functions, to utilize abstractions and approximations, for example as \emph{value function approximation}, state space abstractions \cite{wang2019measuring}, and hierarchical decompositions, cf. \cite{van2009logic}. Many current \emph{deep} RL algorithms too assume the environment behaves as an MDP \cite{wang2020deep}.

To scale to larger problems, the Markov property is no longer adequate, and one may require dependence on a \emph{history} of events and observations. For example, consider a robotic waiter working in a restaurant. It needs to deliver food and beverages to tables, but only \textit{after} it has been requested by guests, and at the end the guests need to pay the price of the items delivered earlier. However, keeping a history of every possible event that ever occurred soon becomes practically infeasible. One well-known class of non-Markov MDP extensions is the \emph{partially observable} MDP \cite{spaan2012partially} in which the current state can be represented as a \emph{probability distribution} over (latent) state features, denoted a \emph{belief state}. Despite the existence of effective POMDP algorithms, many in robotics domains, the general class of POMDPs is computationally much more complex than MDPs, it is not easy to decide what the belief state should include exactly, and how much history should be included, and updating and interpreting the belief states is non-trivial.

A prominent RL direction \cite{liao2020survey} is to model dependence on the arbitrary past \emph{explicitly} resulting in non-Markovian variants of MDPs. Inspired by seminal work \cite{10.5555/1864519.1864559} the idea is to utilize modern logical languages such as \emph{linear temporal logic} \cite{10.1109/SFCS.1977.32} to represent goals and reward functions over past traces, and to employ formal computer science techniques (e.g. \emph{automata}, \emph{verification} and \emph{model-checking}) in decision making. A core idea here is to \emph{compile} a temporal specification of a reward function into an automaton that \emph{monitors} the fulfillment of the temporal formula. Monitors allow for \emph{compiling} the original non-Markov problem back into the MDP framework such that all existing algorithms, including deep RL, can be employed. This fruitful marriage of RL and formal verification combines flexible behavior learning algorithms with formal performance guarantees.

One motivation for employing temporal logic in RL comes from the ability to elegantly specify complex reward structures as in the waiter example, where earnings depend on an ordered series of events in the history. Another, more general, motivation is the need to \emph{constrain} RL behaviors using (declarative) knowledge about which behaviors are desired or considered \emph{safe} \cite{garcia2015comprehensive}, for example to teach an autonomous car how to drive while still obeying traffic rules. Transparent safety of learned behaviors is often part of a general desire for AI systems to behave \emph{responsibly} and \emph{explainable} \cite{van2018ethics,liao2020ethics,kasenberg2020generating}.

In this paper we empirically investigate algorithmic variations in one of the most recently introduced models, \emph{regular decision processes} (RDP) \cite{ijcai2019-766}, in which reward functions \emph{and} transition functions can be specified using temporal logic. We employ RDPs specifically for \emph{grid worlds}, which are archetypical problem scenarios in RL and allow for focused experimentation with new representations and algorithms. More specifically, our contributions are i) a novel \emph{tool chain} implementing RDPs, utilizing exisiting algorithms and tools for RL and model checking, ii) an empirical investigation of the recently introduced RDPs in grid worlds, iii) algorithmic RL extensions to learn RDP behaviors based on Monte Carlo value estimation and incremental (online) compilation of RDPs, and iv) initial steps towards an (empirical) investigation of the trade-offs between temporal logical specifications and the complexity of learning. The paper is organized as follows: we first provide all necessary background in the next section, after which we discuss our approach in \cref{section:approach}, then we continue with an extensive experimental evaluation in \cref{section:experiments} and we conclude in \cref{section:conclusions}.

%
%
\section{Background}
\label{section:background}
Here we will formalize MDPs and basic solution algorithms, after which we introduce non-Markov reward functions and their corresponding temporal logic formalizations. Furthermore we introduce the general compilation of logical specifications into automata functioning as monitors that can be combined with the original MDP into \emph{extended} MDPs, which can be solved using off-the-shelf solution methods. In addition, we describe automata-based \emph{shaping} techniques to deal with the resulting sparse MDPs. Last we introduce RDPs, which support non-Markovian aspects in both reward and transition functions.

%
%
\subsection{Markov Decision Processes}
An MDP $M$ is a tuple $M = \langle S, A, T, R \rangle$, where $S$ is the set of states, $A$ the set of actions, $T: S \times A \times S \rightarrow [0, 1]$ the \emph{transition function} yielding a transition probability and \RewardFunction{} the real-valued \emph{reward function}. Actions only applicable in state $s$ are denoted $a \in A(S)$. A \emph{policy} maps to each state $s \in S$ an action $a \in A$ and is denoted $\pi$.  Additionally, a \emph{discount factor} $\gamma \in [0, 1]$ is used to discount rewards obtained in the future.

As said, MDPs adhere to the \textit{Markov Property}: given the present ($s_{t}$), the future ($s_{t+1}$) is independent on the past ($s_{t-1}$). In other words, everything that is needed to learn from the past is \textit{embedded} in the present state $s_{t}$. The Markov Property holds for all states $s \in S$ and is formally expressed as:
\begin{equation}\label{eq:markov-property}
\nonumber
    p(s_{t + 1} | s_t) = p(s_{t + 1} | s_{1}, s_{2}, \ldots, s_{t})
\end{equation}
\noindent A labelling function \LabellingFunction{}, where $P$ is a finite set of atomic propositions and $S$ the set of states enables a state representation using  \emph{features}.

%
%

\textit{Solving} an MDP comprises computing an optimal policy. A policy is optimal iff it maximizes the expected discounted sum of rewards for every state $s \in S$. Methods for solving decision making problems are generally divided into \textit{model-based} and \textit{model-free} methods \cite{10.5555/3312046}. Model-based methods, generally called \emph{dynamic programming} (DP), can employ the full model ($T$ and $R$) to \emph{plan} optimal sequences of actions. Model-free methods, generally called \emph{reinforcement learning} (RL), do not have knowledge of the model and require sampling, i.e. trial-and-error learning and use that experience to find optimal policies. 

\emph{Dynamic Programming} (DP) methods such as \textit{value and policy iteration} find optimal policies typically by employing a \emph{value function} that expresses for each state \textit{how good} is it for the agent to be in that particular state, and it represents the (expected) discounted future reward that can be obtained from that state, by employing a particular policy. The equation used to calculate a state value is known as the \textit{Bellman Equation}, which formalizes how a state's value, denoted $\upsilon(s)$, is evaluated in terms of expected returns, expressing a relationship between the value of a state and the values of its successor states. DP algorithms use it \emph{iteratively} to update the value of all states until convergence to the \emph{optimal value function} $\upsilon^*(s)$ using the following \textit{Bellman Optimality Equation}:
\begin{equation}\label{eq:bellman-optimality-state}
\nonumber
    \upsilon^{*}(s) = \underset{a \in A}{\text{max}}\;\sum_{s^{\prime} \in S} T(s^{\prime} | s, a)\left[R(s, a, s^{\prime}) + \gamma\;\upsilon^{*}(s^{\prime})\right]
\end{equation}
An optimal action $a$ for $s$ is computed using $\upsilon^*(s)$, $T$ and $R$. 

Where DP methods are concerned with \textit{computing} a value function, RL tries to \textit{learn} value functions using returns obtained from \emph{interaction} with the MDP. In order to find a policy in absence of a model, one needs the \textit{state-action value} for each action $a \in A$ in state $s \in S$, denoted $q(s, a)$, in order to determine the best policy. A straightforward extension of the previous update rule results in $q^{*}(s,a) = \sum_{s^{\prime} \in S} T(s^{\prime} | s, a)\left[R(s, a, s^{\prime}) + \gamma\;\underset{a^{\prime} \in A}{\text{max}}\;q^{*}(s^{\prime},a^{\prime})\right]$. \emph{One-step} RL algorithms employ it to \emph{update} action-values after each step in the environment and select their actions based on $\pi^*(s)=\arg\max_a q(s,a)$.

In addition to \emph{bootstrapping} methods above, where values of states (and actions) are computed using other values, one can employ more unbiased estimation methods for model-free RL such as \emph{Monte Carlo estimation} (MC) in which a value is estimated based on the average return of full sample traces in the MDP, cf. \cite{10.5555/3312046}. In \cref{section:approach} we employ MC as our model-free RL algorithm for RDPs.

%
%
\subsection{Non-Markovian Decision Processes}
If rewards depend on more than just the current state, we end up with Non-Markovian Reward Decision Processes (NMRDPs) \cite{10.5555/1864519.1864559}, a subset of Non-Markovian Decision Processes (NMDPs). Temporal logic can be used to specify the conditions under which reward is obtained. As with MDPs, the states of an NMRDP can be enhanced by labelling function \LabellingFunction{} and propositions $P$, where each state $s \in S$ is a valuation over $P$, thus $s \in 2^{P}$.

Formally, an NMRDP is denoted as the tuple $M = \langle S, A, T, \bar{R} \rangle$, where $S$, $A$ and $T$ are as in an MDP, and $\bar{R}$ is defined as $\bar{R}: (S \times A)^{*} \rightarrow \mathbb{R}$. In words, the reward is specified as a real-valued function over finite state-action sequences, or \textit{traces}, where a trace captures the history of states and is denoted $h = \langle s_{0}, \ldots, s_{k}\rangle$. Because the reward is now dependent on the full history, it no longer fits to define state or state-action values as before. Instead, a temporally extended reward function for a given trace $h$ and reward formulae $\varphi$ is \cite{AAAI1817342}: 
\begin{equation}\label{eq:temporally-extended-reward-function}
    \bar{R}(h) = \sum_{1\;\leq\;i\;\leq\;n\;:\;h\;\models\;\varphi_{i}}r_{i}
\end{equation}
\noindent where the set of pairs $\{(\varphi_{i}, r_{i})_{i=1}^{n}\}$ is assumed to be specified for $\bar{R}$. That is, an agent receives reward $r_{i}$ at state $s \in S$ of trace $h$ that satisfies temporal formula $\varphi_{i}$. The value of a trace $h$ is in turn defined as the accumulation of rewards obtained during trace traversal, possibly discounted by discount factor $\gamma$ \cite{AAAI1817342}. The value of such a trace can now formally be defined as follows: 
\begin{equation}\label{eq:trace-value}
\nonumber
    \upsilon(h) = \sum_{k = 1}^{|h|}\gamma^{k - 1}\;\bar{R}(\langle h(1), h(2), \ldots, h(k)\rangle)
\end{equation}
\noindent where discount factor $\gamma \in [0, 1]$ as usual and $h(k)$ denotes the pair $(s_{k - 1}, a_{k})$. Because NMRDPs define the value of traces instead of individual states, a policy no longer maps states to actions as before. Instead, a policy for an NMRDP is a mapping from histories to actions. The value of a policy in terms of expected return thus becomes the expected discounted sum of rewards over a possibly infinite amount of traces. The distribution over traces is defined by the initial state $s_{0}$, the transition function $T$ and policy $\pi$. The expected value of infinite traces can formally be defined as $\upsilon_{\pi}(s) = E_{h\;\sim\;M,\pi,s_{0}}\upsilon(h)$.

%
%
\subsection{Temporal Logic, Automata and Product MDPs}
\label{section:ldl}
Temporal logic to express non-Markovian aspects has a history \cite{10.5555/1864519.1864559,10.1109/SFCS.1977.32} containing, e.g., Linear Temporal Logic \cite{10.1109/SFCS.1977.32} (LTL). It uses the standard Boolean connectives of propositional logic, i.e. $\land$, $\lor$ and $\lnot$, with the addition of temporal connectives $G$ (\textit{always}), $F$ (\textit{eventually}), $X$ (\textit{next}) and $U$ (\textit{until}). More recent variations restrict to finite traces: \textit{Linear Temporal Logic over Finite Traces}, denoted \LTLf{}, and \textit{Linear Dynamic Logic over Finite Traces} which allows for regular expressivity \cite{10.5555/2540128.2540252}. Using \LDLf{}, goals can be as expressive as regular expressions while at the same time providing a more attractive specification syntax. Formally, \LDLf{} formulae $\phi$ can be built using an atomic property $\mathit{tt}$ for the logical $\mathit{true}$, a propositional formula $\varphi$ and a path expression $\rho$, which is a regular expression over propositional formulae $\phi$. In addition to regular expression constructs, $\rho$ uses a test construct $\varphi?$, indicating to only continue evaluation when $\varphi$ evaluates to $\mathit{true}$. The \LDLf{} formalism, as presented by \cite{10.5555/2540128.2540252}, is expressed in \cref{eq:ldlf-phi,eq:ldlf-rho}.

\begin{equation}\label{eq:ldlf-phi}
    \varphi ::= \mathit{tt}\;|\;\lnot\varphi\;|\;\varphi_1 \land \varphi_2\;|\;\langle\varrho\rangle\varphi
\end{equation}
\begin{equation}\label{eq:ldlf-rho}
    \varrho ::= \phi\;|\; \varphi?\;|\;\varrho_1 + \varrho_2\;|\;\varrho_1 ; \varrho_2\;|\;\varrho^{*}
\end{equation}

\noindent Intuitively, one may interpret \LDLf{} formula $\langle\varrho\rangle\varphi$ as stating that, from the current step in the trace, there exists \textit{at least one} (cf. $\exists$) execution path that satisfies regular expression $\varrho$ such that the last step in the trace satisfies $\varphi$. Conversely, $[\varrho]\varphi$ states that, from the current step in the trace, \textit{all} (cf. $\forall$) execution paths satisfying regular expression $\varrho$ are such that the last step in that execution path satisfies $\varphi$. For example, to formalize the property of a robotic waiter to always serve guests after they have placed an order, the formula $[\mathit{true}*](\mathit{order} \rightarrow \langle\mathit{true}^{*};\mathit{served}\rangle)\mathit{end}$ can be used.

Temporal formulae specified using \LDLf{} can be compiled into Deterministic Finite Automata (DFA) \cite{AAAI1817342}. Formally, a DFA for formula $\varphi$ is denoted \DFA{}, where $2^{P}$ is the input alphabet containing all truth assignments to propositions in $P$, $Q$ is the state space, $\delta$ the transition function, $F$ the set of accepting states and $q_{0}$ the initial state.

\begin{figure}[t]
    \centering
    \subfloat[Safety]{
        \includegraphics[width=2cm]{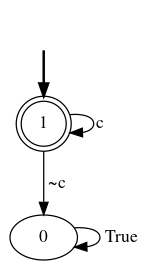}
        \label{fig:automaton-property-safety}
    }
    \subfloat[Liveness]{
        \includegraphics[width=2cm]{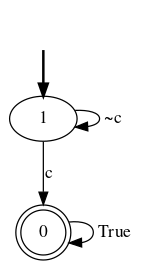}
        \label{fig:automaton-property-liveness}
    }
    \subfloat[Grid World]{
    \scalebox{.65}{\input{images/grid-world-rdp-transition}}
    \label{fig:grid-world:rdp}
    }
    \hfill
    \caption{(left) Automata for \LDLf{} formulae: a) $[\mathit{true}^{*}]\langle c^{*} \rangle \mathit{end}$, b) $\langle\mathit{true}^{*}; c;\mathit{true}^{*} \rangle \mathit{end}$ (right) A grid world modeled as an RDP}
    \label{fig:ldf-formulae-automata}
\end{figure}

Core properties that can be expressed in \LDLf{} are \textit{safety} and \textit{liveness} \cite{10.5555/2540128.2540252}. A safety property is used to indicate that \textit{something bad should never happen}, or \textit{something good always holds}, and can be expressed as $[\mathit{true}^{*}]\langle c^{*} \rangle \mathit{end}$, where $c$ indicates the good condition and the asterisk (*) indicates $c$ holds at every step up until and including the last step of the trace. That is, \textit{until the end of the trace}, $c$ holds. Conversely, a liveness property indicates that some condition should be met \textit{before the end of the trace} and can be expressed as $\langle\mathit{true}^{*}; c;\mathit{true}^{*} \rangle \mathit{end}$, where $c$ is the condition to be met. In words, \textit{eventually before the end of the trace}, $c$ holds. \Cref{fig:automaton-property-safety,fig:automaton-property-liveness} visualize this.

\noindent \textit{Solving} an NMRDP $M = \langle S, A, T, \{(\varphi_{i}, r_{i})_{i=1}^{m}\} \rangle$, with temporal formulae $\varphi_{i}$ and $r_i$ the corresponding rewards, is tackled by formulating the \emph{extended} MDP $M^{\prime}$ as $M^{\prime} = \langle S^{\prime}, A, T^{\prime}, R^{\prime} \rangle$ that is \textit{equivalent} to $M$ in the sense that states can be mapped in such a way that the mapping yields identical transition probabilities for $T$ and $T^{\prime}$. Each formula $\varphi_{i}$ is compiled into an equivalent automaton, as in \cref{fig:automaton-property-safety,fig:automaton-property-liveness}, and the \emph{cross-product} between the original NMRDP $M$ and these automata is computed, resulting in the extended MDP $M^{\prime}$. Some straightforward choices should still be made about discounting to prevent infinite reward exploitation and whether rewards belonging to a formula $\varphi$ can be obtained only once or multiple times. We omit formal details of this standard construction (but cf. \cite{AAAI1817342,nlenaers}) and refer here to an example later in this paper: \cref{fig:exp-safety} shows a grid world MDP where a red square needs to be avoided, something which is specified using the \LDLf{} formula $\varphi \equiv [\mathit{true}^{*}]\langle(\lnot x_{is1} \land \lnot y_{is2})^{*} \rangle\mathit{end}$, and where the extended MDP depicted in~\cref{fig:exp-safety:product-mdp} is the result of the cross product between the automaton representing $\varphi$, the grid world MDP, and the automaton representing an additional formula expressing a reward of $+50$ when reaching the top right corner. Note that the extended model is again an MDP where typical RL and DP algorithms can be employed.

%
%
\subsection{Regular Decision Processes: Non-Markovian Dynamics}
\label{sec:rdp}
The concept of an NMRDP can be extended to a decision process in which not only the reward function, but the transition function too can depend on the arbitrary past, and where both are represented using a logic like \LDLf{} As described in \cref{section:ldl}, these, in turn, can be compiled into automata, which allows for rewards and transitions to be \emph{monitored}, and compiled into product models yielding an MDP. Such non-Markovian transitions were introduced in regular decision processes (RDP) \cite{ijcai2019-766}, which is a fully observable, probabilistic, non-Markovian, sequential decision making model, where successor states and rewards can be stochastic functions of the entire history. Just like before, RDP states are endowed with a labeling function over a set of predicates.

An RDP $M$ is defined as the tuple $M = \langle P, S, A, \mathit{Tr}_{L}, R_{L}, s_{0}\rangle$, where $P$ is the set of propositions that induces state-space $S$ with initial state $s_{0}$, $A$ the set of actions, $\mathit{Tr}_{L}$ the transition function and $R_{L}$ the reward function, where both $\mathit{Tr}_{L}$ and $R_{L}$ are now non-Markovian. Transition function $\mathit{Tr}_{L}$ is defined by a finite set $T$ of quadruples of the form $(\varphi, a, P^{\prime}, \pi(P^{\prime}))$, where $\varphi$ is an \LDLf{} formula over $P$, $a \in A$ an action, $P^{\prime} \subseteq P$ the set of propositions $p \in P$ that are affected by $a$ when $\varphi$ holds and $\pi(P^{\prime})$ the distribution over proposition in $P^{\prime}$ that describe the post-action distribution. The reward function $R_{L}$ is specified using a finite set $R$ of pairs $(\varphi, r)$, where $\varphi$ is an \LDLf{} formula over propositions in $P$ and $r \in \mathbb{R}$ a real-valued reward. It is assumed that for the quadruples in $T$, the value of variables not in $P^{\prime}$ are not affected by action $a$ \cite{ijcai2019-766}. If the set \RDPTransitionQuadruples{} defines all quadruples for $a$, then all formulae $\varphi_{i}$ must be \textit{mutually exclusive} such that $\varphi_{i} \land \varphi_{j}$ is inconsistent for $i \neq j$. In other words, no two formulae $\varphi_{i}$ and $\varphi_{j}$ can hold at once if both apply to action $a$ and $\varphi_{i}$ and $\varphi_{j}$ are not identical. In addition, let $s^{\prime}|_{P^{\prime}}$ denote the restriction of $s^{\prime}$ to properties in $P^{\prime}$. Then, $\mathit{Tr}_{L}$ is defined as $\mathit{Tr}_{L}((s_{0}, \ldots, s_{k}), a, s^{\prime}) = \pi(s^{\prime}|_{P^{\prime}})$ if quadruple $(\varphi, a, P^{\prime}, \pi(P^{\prime}))$ exists such that $s_{0}, \ldots, s_{k} \models \varphi$ and $s_{k}$ and $s^{\prime}$ agree on all variables in $P \setminus P^{\prime}$. That is, given trace $s_{0}, \ldots, s_{k}$, action $a$ and quadruple $(\varphi, a, P^{\prime}, \pi(P^{\prime}))$ with formula $\varphi$ that is satisfied by $s_{0}, \ldots, s_{k}$, $s^{\prime}$ is a possible next state if it assigns the same value to all propositions not in $P^{\prime}$. If this is the case, then the transition probability equals the probability $\pi$ assigns to $s^{\prime}|_{P^{\prime}}$. In all other cases, $\mathit{Tr}_{L}((s_{0}, \ldots, s_{k}), a, s^{\prime}) = 0$.

As an illustration, consider \cref{fig:grid-world:rdp}, outlining a $3 \times 3$ grid world with the upper-left state $s_{11}$ being the initial state and the upper-right state $s_{31}$ being a terminal state. Let us define a transition that intuitively states that, when an agent goes east in the bottom-left state $s_{13}$ and ends up in the bottom-center state $s_{23}$, immediately followed by going east \textit{again} in $s_{23}$, the probability of ending up in the bottom-right state $s_{33}$ is set to $0.1$, denoted $\pi(s_{33} | \{x_{is2}, x_{is3}\}) = 0.1$. Otherwise, $Tr(s_{23}, e, s_{33}) = 1$. In other words, the transition from $s_{23}$ to $s_{33}$ depends on the transition from $s_{13}$ to $s_{23}$. In addition, the propositions affected by this transition are defined by $P^{\prime}$ such that $P^{\prime} \subseteq P = \{x_{is2}, x_{is3}\}$. All other propositions are not affected by said transition. Both transitions can be captured by \LDLf{} formula $\varphi_{1}$ and $\varphi_{2}$ as $\varphi_{1} = \langle \mathit{true}^{*};\lnot x_{is1} \lor \lnot y_{is3};x_{is2} \land y_{is3} \rangle\mathit{end}$ and $\varphi_{2} = \langle \mathit{true}^{*};x_{is1} \land y_{is3};x_{is2} \land y_{is3} \rangle\mathit{end}$. Given $\varphi_{1}$ and $\varphi_{2}$, we can define a quadruple for $e$ that uses $\varphi_{1}$ or $\varphi_{2}$ respectively as $(\varphi_{1}, \{x_{is3} \land y_{is3}\}, 1)$ and $(\varphi_{2}, e, \{x_{is3} \land y_{is3}\}, 0.1)$. For brevity, we assume these are the only quadruples for $e$, conforming to exhaustiveness and mutual exclusion \cite{ijcai2019-766}. Then, let us define two traces $h_{1}$ and $h_{2}$ that each reach $s_{33}$ differently as $h_{1} = \langle s_{11}, s_{12}, s_{13}, s_{23}, s_{33} \rangle$ and $h_{2} = \langle s_{11}, s_{12}, s_{22}, s_{23}, s_{33} \rangle$. Then, using the aforementioned quadruple for $e$, the affected propositions $P^{\prime}$ and the definition of $Tr_{L}$, i.e. $\mathit{Tr}_{L}((s_{0}, \ldots, s_{k}), a, s^{\prime}) = \pi(s^{\prime}|_{P^{\prime}})$, the transition functions on $h_{1}$ and $h_{2}$ from $s_{23}$ to $s_{33}$ can be calculated as $Tr_{L}(h_{1}, e, s_{33}) = \pi(s_{33} | \{x_{is2}, x_{is3}\}) = 1$ and $Tr_{L}(h_{2}, e, s_{33}) = \pi(s_{33} | \{x_{is2}, x_{is3}\}) = 0.1$.

Solving an RDP involves the well known construction of an \textit{extended} MDP as a product of all automata monitoring the satisfaction of (transition and reward) \LDLf{} formulae combined with the initial RDP state space \cite{AAAI1817342,10.5555/2540128.2540252}, resulting in an MDP that can again be solved by off-the-shelf algorithms. Note that, because of the combinatorial nature of this construction, the extended MDP does not necessarily scale well. The equivalence between the RDP and the constructed MDP entails that optimal policies found in the constructed MDP can be mapped back to the RDP, thus yielding optimal policies for the initial RDP.

The product models employed in non-Markovian decision process solutions grow quickly with the number of formulas, see the example in \cref{section:ldl}. The result of non-Markovian dependencies is that paths to receiving rewards can become long, and complicate typical bootstrapping RL methods and exploration. One general solution for MDPs is \emph{reward shaping} \cite{10.5555/645528.657613} (RS): giving intermediate rewards to speed up learning, with the restriction that the extra rewards do not alter the optimal policy. So-called \emph{potential-based} RS replaces the original reward function \RewardFunction{} by an alternative reward function \RewardShapingFunction{}, where \ShapingRewardFunction{} is a \textit{shaping reward function}. In turn, this function can be applied to \textit{potential-based} RS of the form \PotentialRewardFunction{}. The way in which RS is applied inherently depends on the representation of the reward function. For NMRDPs an opportunity arises to utilize the structure of the DFA representing a reward function \cite{RG326459658}. Every step in the extended MDP can be given a reward proportional to the \emph{distance} in that DFA to an accepting state (i.e. when the original reward would be given).

%
%

\subsection{Related Work}
The typical MDP context is well studied and there is an abundance of algorithms and representations \cite{van2009logic,wiering2012reinforcement,10.1002/9780470316887,10.5555/3312046}. Endowing MDPs with non-Markovian goal and reward functions has a history with seminal work on model-based settings \cite{10.5555/1864519.1864559,thiebaux2006decision} and more recently several subclasses are considered (e.g. probabilistic vs. deterministic) \cite{AAAI1817342,brafman2019planning}. The most recent addition to the field are the general \emph{regular decision processes} \cite{ijcai2019-766} we employ here. One aim of all these methods is to scale MDPs to more complex problems. However, another main reason to utilize temporal logics for reward specifications is that it opens up many new possibilities for \emph{reward function engineering}, resulting in more intuitive and technically useful ways to specify tasks and goals. A more general view, based on automata as \emph{transducers} \cite{KR2020-89} improves on the technical part by merging the non-Markovian parts into a single structure.

The use of temporal logics \cite{10.1109/SFCS.1977.32,10.5555/2540128.2540252} in model-free RL settings is a recent trend \cite{liao2020survey}, and comes with additional requirements since the model of the environment is unknown. Many ideas here come from \emph{constrained} or \emph{safe} \cite{garcia2015comprehensive} forms of RL, where the policy space is restricted either before learning, or during action selection, based on a notion what are (un)desired actions. Safety issues have obvious connections with \emph{model checking} \cite{giaquinta2018strategy} and some recent RL approaches instantiate that connection for safe RL \cite{alshiekh2017safe,fulton2018safe}. Very recently, several approaches have appeared combining formal temporal logic with RL \cite{li2017reinforcement,AAAI1817342,RG326459658,camacho2019ltl,de2019foundations,KR2020-89}. Some focus more on the representational devices such as \emph{reward machines} \cite{camacho2019ltl}, some study additional mechanisms such as \emph{shaping} to aid in the more complex learning process \cite{10.5555/645528.657613}, and others introduce variants such as \emph{geometric} LTL to capture a different semantics of goals \cite{littman2017environment}. Overall, variations exist in different logics, different underlying automata (e.g. DFA vs Mealy) and inference algorithms, and different RL algorithms to solve the resulting extended MDPs.

The meaning of ''model-free'' has variations here, since one can can assume that nothing is known, or that at least the reward formula is (which is quite believable when we want an agent to adhere to certain rules or restrictions). In the latter case one can use the monitor automata states as extra state information and apply any form of deep function approximation \cite{jothimurugan2020composable}. In general, reward and transition functions may need to be learned from traces for fully general RL systems. In the temporal logic settings we describe, this typically amounts to \emph{automaton induction} algorithms that can work on examples of traces (positive or negative) in deterministic or even probabilistic settings, which contains notoriously hard settings, but some promising work is emerging \cite{camacho2019learning,kim2019bayesian,furelos2021induction}. In the context of RDPs, initial work with a Mealy machine representation shows promise \cite{abadi2020learning}. In addition, temporal logic allows for declarative and intuitive models, hence in terms of explainability in RL many possibilities are left, and only some work is just emerging \cite{kasenberg2020generating}.

%
%
\section{Approach and Software Design}\label{section:approach}
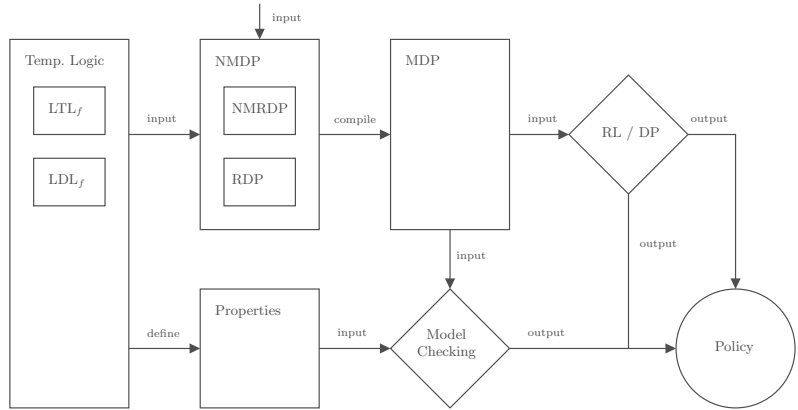
\begin{figure}[t]
    \centering
    \scalebox{0.60}{\input{images/project-overview}}
    \caption{Conceptual/Tool chains}
    \label{fig:project-overview}
\end{figure}

In this paper we develop a new \emph{tool chain} for the recently introduced RDPs and experiment with algorithmic variations, specifically applied to grid worlds (cf. \cite{nlenaers}). \cref{fig:project-overview} graphically shows a simplified high-level overview of how decision processes, temporal logic and model checking intertwine. Currently no software tool can conveniently model, visualize and solve all RDPs, which motivates our particular approach. Secondly, RDPs are introduced very recently and not much empirical evidence has been gathered so far \cite{abadi2020learning,RG326459658}. Also the familiar grid worlds in general are underrepresented in the temporal logical RL community despite their abundance in basic RL research, and despite their ability to quickly show insight into models.
In general we follow the main paths through \cref{fig:project-overview}, where rectangles, diamonds and circles represent formalisms (or models), processes and artifacts, respectively. On the left we see temporal logics such as \LTLf{} / \LDLf{} used to define non-Markov decision models, as we have seen in the previous sections, where we also described how these can be compiled into (extended) MDPs, which can then be solved by traditional MDP algorithms. Note that NMDPs are not solely dependent on temporal logic, but require other input such as a state space definition. In the lower-most flow, temporal formulae can be used to define system properties that can be verified using \emph{model checking}~\footnote{In the current paper there is no room to highlight it, but model-checkers such as Storm (\url{https://www.stormchecker.org/}) can be employed for shaping and shielding purposes (and more) in this tool chain, cf. \cite{nlenaers}).}. Finally, experimental learning algorithms can be combined with formal verification methods to produce a policy.

Our prototype integrates existing software tools. First, an integration is made with FLLOAT \cite{ijcai2019-766}, a tool that allows to construct automata from \LTLf{} and \LDLf{} formulae. Because the prototype is a TypeScript (TS) web application, and FLLOAT is built with Python, a small web server is put in place to communicate with FLLOAT. Communication then occurs by making HTTP requests from the prototype through a \textit{Browser HTTP Layer} to the FLLOAT application through a \textit{Server API Layer}. In addition, an integration with a browser-based Graphviz extension called Viz.js\footnote{https://github.com/mdaines/viz.js} was made. It allows for \emph{visualization} of automata within the prototype. Input to the software prototype is defined in terms of TS variables, stored in a single TS file an presented during execution at runtime.

%
%
\subsection{Compilation: from RDP to MDP}
A core component in our approach is the conversion of NMDPs to MDPs for both \textit{off-line}, i.e. before learning, and \textit{on-line}, i.e. during learning, use cases. Intuitively, one can think of the off-line case as a model-based control problem, where the reward function and transition function are fully known to the agent. However, in contrast to other work, we compute the extended MDP \emph{incrementally}. On the other hand, the on-line case can be thought of as a model-free control problem where the agent has to interact with the environment to learn an optimal behavior. Also here the algorithm constructs the extended MDP incrementally, but now only in the areas of the state-action space that are actually experienced by the agent in the interaction with the environment. In addition, in this model-free setting it is assumed that the agent has access to only the states of the automata tracking the formulae, just like in other works (e.g. \cite{jothimurugan2020composable}). Throughout the algorithms, automata for rewards are indicated by $Q^{R}_{i}$ and automata for transitions are indicated by $Q^{T}_{j}$ and their union is denoted $Q_{k}$. 

%
%
The compilation of an NMDP can exploit knowledge of the known dynamics/reward model. \cref{alg:nmdp-to-mdp-off-line} outlines our algorithm, generalized to NMDPs. It incrementally builds an extended MDP \textit{off-line} by incorporating all \LDLf{} automata such that only reachable states are generated. Here, off-line means the compilation is done before solving the final MDP.
\begin{algorithm}[t]
    \caption{NMDP to MDP (off-line)}
    \label{alg:nmdp-to-mdp-off-line}
    \input{algorithms/nmdp-to-mdp-off-line}
\end{algorithm}
\noindent The transition function in \cref{alg:nmdp-to-mdp-off-line} on \cref{line:nmdp-to-mdp-off-line:transition} abstracts away the different transition dynamics between NMRDPs and RDPs by using labelling function $\Lb$, making it applicable to both models. Furthermore, the state space generated by \cref{alg:nmdp-to-mdp-off-line} is \textit{minimal} because it only generates states that are reachable, and thus solution algorithms do not waste time on irrelevant states. The resulting MDP can be solved using e.g. \emph{value iteration}, cf.~\cite{nlenaers}.

Because in the model-free setting the reward function and transition dynamics are not known a priori, compilation cannot occur in a similar fashion as in \cref{alg:nmdp-to-mdp-off-line}. We employ a different, \emph{online, incremental} approach in \cref{alg:nmdp-to-mdp-on-line}. Similar to the off-line algorithm it \emph{incrementally} builds the extended MDP, only here the automata $Q^{R}_{i}$ for rewards and automata $Q^{T}_{j}$ for transitions are \emph{not} known to the agent. Hence, $Q_{k}$ is not defined as input like it is for \cref{alg:nmdp-to-mdp-off-line}. Furthermore, the extended MDP is not fully defined in terms of dynamics of transitions and rewards. This, in turn, requires an environment capable of handling \LDLf{} automata for rewards and transitions. In addition to \cref{alg:nmdp-to-mdp-on-line}, a step function first gets the current state $s_{t}$ from the environment using $s_{t} \leftarrow \mathtt{env.snapshot()}$. In addition, all automata are retrieved through $Q_{k, t} \leftarrow \mathtt{env.get\_automata\_states()}$. Then, for each $q_{k, t} \in Q_{k, t}$, both the original state and all automata states transition to their subsequent states through $s_{t + 1} \leftarrow \mathtt{env.transition}(s_{t}, a)$ and $q_{k, t + 1} \leftarrow \mathtt{transition}(q_{i, t}, \Lb(s_{t + 1}))$ respectively. Automata are then updated through $\mathtt{env.set\_automaton\_state}(Q_{k}, q_{k, t + 1})$ the reflect the state transition. Finally, when $Q_{k, t}$ has been iterated over, i.e. all automata have transitioned, a next state is generated by $s^{\prime}_{t+1} \leftarrow (q_{1,t + 1}, q_{2,t + 1}, \allowbreak \ldots, q_{k,t + 1}, s_{t + 1})$, i.e. the MDP state is extended with each \emph{monitor} state. In addition, the rewards for all automata currently in an accepting state are summed by $r \leftarrow \mathtt{sum\_accept}(Q_{k, t + 1} \setminus Q^{T}_{i, t + 1})$. Indeed, the better part of \cref{alg:nmdp-to-mdp-on-line} aligns with first-visit MC \cite{10.5555/3312046}, except that the underlying problem definition is assumed to be non-Markovian and hence compiled \textit{on-line} from NMDP to MDP.
\begin{algorithm}[t]
    \caption{NMDP to MDP (on-line)}
    \label{alg:nmdp-to-mdp-on-line}
    \input{algorithms/nmdp-to-mdp-on-line}
\end{algorithm}

Similar to \cref{alg:nmdp-to-mdp-off-line}, \cref{alg:nmdp-to-mdp-on-line} generate only reachable states and is therefore \textit{minimal}. This is due to the transition function of automata being defined as $q_{k, t + 1} \leftarrow \mathtt{transition}(q_{i, t}, \Lb(s_{t + 1}))$, where a transition cannot occur if the target state is unreachable. Due to the nature of RL, the implicitly extended MDP contains only states actually encountered by an agent through interaction with the environment. Note that, as opposed to \cref{alg:nmdp-to-mdp-off-line}, \cref{alg:nmdp-to-mdp-on-line} does \textit{not} contain all information on the history of states per se. Due to the trial-and-error nature of MC, some states might remain unobserved after \cref{alg:nmdp-to-mdp-on-line} has completed. Therefore, an optimal policy $\pi^{*}$ is only guaranteed in the limit.

%
%
\section{Experiments}\label{section:experiments}
Our experimental evaluation focuses on RDPs for grid worlds, utilizing a model-free online MC algorithm. Our experimental evaluation focuses on RDPs for grid worlds, utilizing a model-free online MC algorithm. Overall, the goal is to empirically assess various aspects of RL for RDPs, with a focus on the relation between RDP elements and final learning performance in the resulting extended MDP. For more experiments, (also model-based , value iteration), cf. \cite{nlenaers}. In this section we target four different empirical questions: \textbf{R1}: \emph{How does learning performance relate to goal sparsity/complexity?}, \textbf{R2}: \emph{How can shaping help for complex goals?}, \textbf{R3}: \emph{What are the implications of safety properties on learning performance?}, and \textbf{R4}: \emph{What is the relation between learning performance and non-Markovian dynamics?}

%
%
\begin{experiment}{goal-sparsity}{Goal Sparsity}
\begin{figure}[t]
        \centering
        \subfloat[Not so sparse]{
            \scalebox{.85}{\input{images/experiment-goal-sparsity-adjacent}}
            \label{fig:exp-goal-sparsity-adjacent}
        }
        \hfill
        \subfloat[Somewhat sparse]{
            \scalebox{.85}{\input{images/experiment-goal-sparsity-center}}
            \label{fig:exp-goal-sparsity-center}
        }
        \hfill
        \subfloat[Very sparse]{
            \scalebox{.85}{\input{images/experiment-goal-sparsity-diagonal}}
            \label{fig:exp-goal-sparsity-diagonal}
        }
        \hfill
        \caption{\cref{experiment:goal-sparsity} - Goals for on-line compilation with RL}
        \label{fig:exp-goal-sparsity-setup}
    \end{figure}
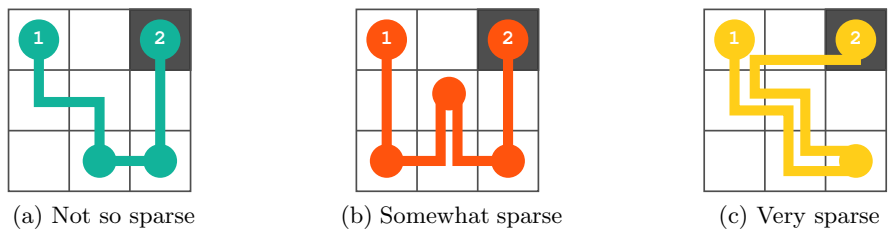
    This experiment aims at relating \textit{goal sparsity} to the performance fist-visit MC. Here, goal sparsity describes the accumulated minimum length of traces $h_{i}$ accepted by \LDLf{} formulae $\varphi_{i}$. The idea is to increase the grid world size, while keeping reward formulae constant, such that the traces increase in length due to an increase in the size of the state space. To illustrate this, temporal formulae encoding liveness properties are used such that the number of steps to satisfy a formula increases with the grid world size. The minimum length of a trace $h_{i}$ is measured in terms of the minimum number of states contained in $h_{i}$ for it to be accepted. The quantitative measurement is defined by the relative frequency of values within $10\%$ of the maximum value. This range is deemed acceptable for a solution as it follows from the value used for $\epsilon$, being $\epsilon = 0.1$, that generates exploration noise. Data is gathered over 50 runs, where each run consists of 1000 episodes with a maximum of 50 steps per episode. Furthermore, $\gamma = 1$. In order to consistently increase the goal sparsity when the grid world size is increased, the agent always starts in state $s_{11}$ and an episode is terminated when the agent reaches $s_{13}$, after which a new episode is initiated until the maximum number of episodes is reached. \Cref{fig:exp-goal-sparsity-setup} outlines three goals, each rewarded $+1000$, with \cref{fig:exp-goal-sparsity-adjacent} being the least sparse where goal state are adjacent, \cref{fig:exp-goal-sparsity-center} being somewhat sparse where goal states require the agent to travel through the center of the grid and \cref{fig:exp-goal-sparsity-diagonal} being the most sparse where the agent is required to go reach the far-right state and then go back to its initial state again. The goals encoded in \LDLf{} as $\langle \mathit{true}^{*};x_{is2} \land y_{is3};\mathit{true}^{*};x_{is3} \land y_{is3};\mathit{true}^{*} \rangle\mathit{end}$, $\langle \mathit{true}^{*};x_{is1} \land y_{is3};\mathit{true}^{*};x_{is2} \land y_{is2};\mathit{true}^{*};x_{is3} \land y_{is3};\mathit{true}^{*} \rangle\mathit{end}$ and $\langle \mathit{true}^{*};x_{is3} \land y_{is3};\mathit{true}^{*};x_{is1} \land y_{is1};\mathit{true}^{*} \rangle\mathit{end}$ respectively.
    
    \Cref{fig:exp-goal-sparsity-results} outlines the results for this experiment. It shows that the latter two goals quickly become harder to solve in this setting, as there is a steep decline in optimal behaviour between grid world sizes $3 \times 3$ and $4 \times 4$. In addition, the graph shows that for the more sparse goals the chance of finding optimal behaviour under the conditions outlined for this setting for a grid world of $7 \times 7$ becomes nil. Thus, the observed data indicates there is a relation between goal sparsity and the performance of first-visit MC.
    
    \begin{figure}[t]
        \centering
        \includegraphics[width=12cm]{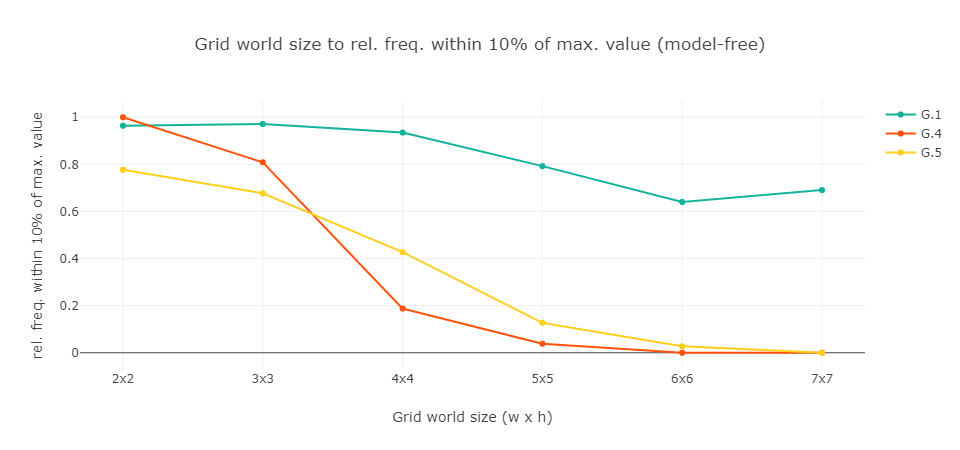}
        \caption{Rel. freq. within 10\% of max. value}
        \label{fig:exp-goal-sparsity-results}
    \end{figure}
\end{experiment}

%
%
\begin{experiment}{reward-shaping}{Reward Shaping}
    Recall that for \cref{experiment:goal-sparsity}, sparser goals quickly become harder to solve. Therefore, this experiment aims to apply RS to an RDP construction from \cref{section:approach} so as to identify whether the performance of MC can be improved when using a potential function from \cref{sec:rdp}. In addition, \cref{alg:nmdp-to-mdp-on-line} will be used for on-line compilation in a model-free setting. A preliminary experiment showed that for a $5 \times 5$ grid world, in which $\langle \mathit{true}^{*};x_{is1} \land y_{is5};\mathit{true}^{*};x_{is3} \land y_{is3};\mathit{true}^{*};x_{is5} \land y_{is5};\mathit{true}^{*} \rangle\mathit{end}$ is used, an optimal policy is rarely found \cite{nlenaers}. Therefore, this experiment outlines the effect of applying a potential function for RS. For this experiment, a $5 \times 5$ grid world is used, transitions are deterministic and MC is applied as the RL learning algorithm. The reward for the goal is set to $+1000$. \Cref{fig:exp-reward-shaping:mf-setup} outlines a possible optimal trace for the given goal. The quantitative measurements are defined by the averaged returns per episode and the size of the extended MDP. A total of 50 runs with each 1000 episodes and a maximum of 50 steps per episode is used. Parameters are defined as $\gamma = 1$ and $\epsilon = 0.1$. The agent always starts an episode in state $s_{11}$ and an episode is terminated when the agent reaches $s_{51}$, after which a new episode is initiated until the maximum number of episodes is reaches.
    
    \begin{figure}[t]
        \centering
        \subfloat[Shaping experiment]{
        \scalebox{.60}{\input{images/experiment-reward-shaping}}
        \quad
        \label{fig:exp-reward-shaping:mf-setup}
        }
        \subfloat[Unsafe]{
            \input{images/experiment-safety-unsafe}
            \quad
            \label{fig:exp-safety:unsafe}
        }
        \subfloat[Safe]{
            \input{images/experiment-safety-safe}
            \label{fig:exp-safety:safe}
        }
        \hfill
        \caption{(a) Possible optimal trace for reward shaping experiment, (b)-(c) Possible optimal traces for safety experiment}
        \label{fig:exp-safety}
    \end{figure}
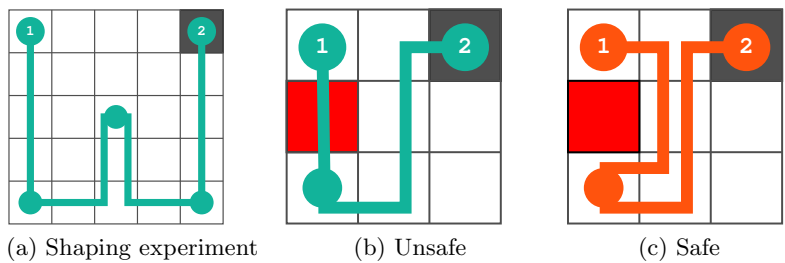
    
    \Cref{fig:exp-rs:mf} plots the results for this experiment. From \cref{fig:exp-reward-shaping-avg-return} it can be observed that a shaped reward takes somewhat longer to learn, but the averaged return is significantly higher for the $5 \times 5$ grid world. More specifically, the averaged return for unshaped rewards shows that unshaped rewards are, on average, not received, as the trend remains just below zero. Finally, \cref{fig:exp-reward-shaping-ext-mdp-size} outlines that shaped rewards reach far more states when compared to unshaped rewards. Given the observation that unshaped rewards are, on average, not received, all states reachable only after a goal is satisfied are very rarely explored for unshaped rewards. Hence, the size of the extended MDP is significantly smaller.
    
    As an intuitive evaluation of the observed results, recall the $5 \times 5$ grid world as outlined in \cref{fig:exp-reward-shaping:mf-setup}. Finding a trace that follows the critical path for the given goal without a potential function is then inherently hard. Consider, for example, the trace outlined by \cref{fig:exp-reward-shaping:mf-setup}. This trace contains $16$ consecutive steps, where the agent may stray from the path at each one of these steps with a probability $\epsilon$. Even if the agent reaches the end of the trace in the case of unshaped rewards, the back-propagation of the reward value may be insignificant when it updates the state-action value of state $s_{11}$, as the agent only gets rewarded for the trace when it finally reaches terminal state $s_{51}$. In turn, on average, the agent might only obtain the unshaped reward relatively rarely, leaving most of the states that account for the latter part of the trace uncharted. Note that, as discussed, this result is accounted for in \cref{fig:exp-reward-shaping-ext-mdp-size}. Conversely, a shaped reward encourages the agent to better follow the critical path, in turn increasing the probability of satisfying the reward formula by its trace and thus increasing the number of explored states in the extended MDP generated by \cref{alg:nmdp-to-mdp-on-line}. In general, it can be observed that shaped rewards make learning perform better, while at the same time increasing the size of the (encountered) state space and the number of steps required in optimal traces.
    
    \begin{figure}[t]
        \centering
        \subfloat[avg. return]{
            \includegraphics[width=5.75cm]{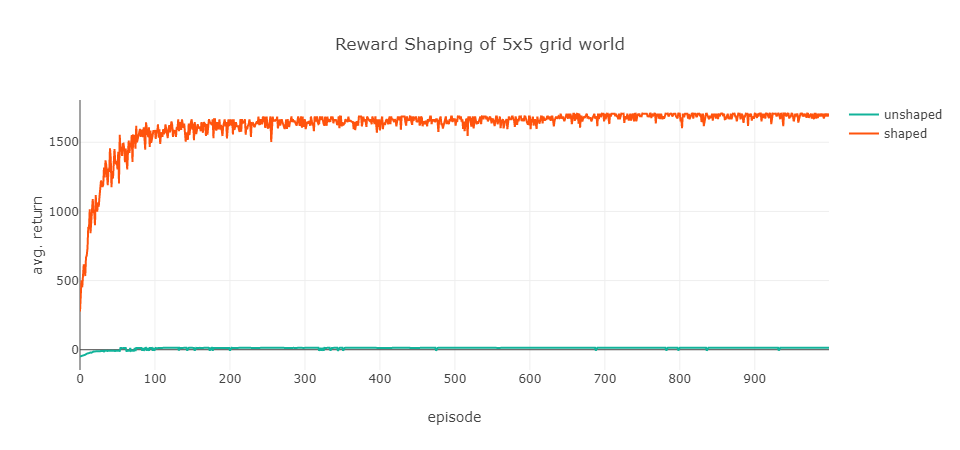}
            \label{fig:exp-reward-shaping-avg-return}
        }
        \subfloat[ext. MDP size]{
            \includegraphics[width=5.75cm]{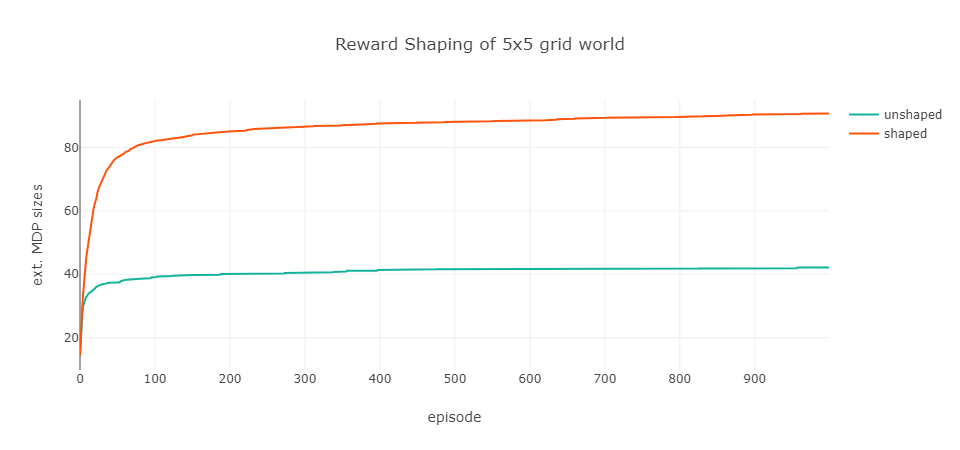}
            \label{fig:exp-reward-shaping-ext-mdp-size}
        }
        \hfill
        \caption{MC performance for unshaped vs. shaped rewards.}
        \label{fig:exp-rs:mf}
    \end{figure}
    
    When reward shaping is applied, a significant increase in MC performance can be observed from \cref{fig:exp-reward-shaping:top-ten-percent-increase} for a $5 \times 5$ grid world. Where the unshaped reward decreases rapidly between $3 \times 3$ and $4 \times 4$ grid world, the shaped reward decreases significantly less over the course of the increasing grid world sizes.
    
    \begin{figure}
        \centering
        \includegraphics[width=12cm]{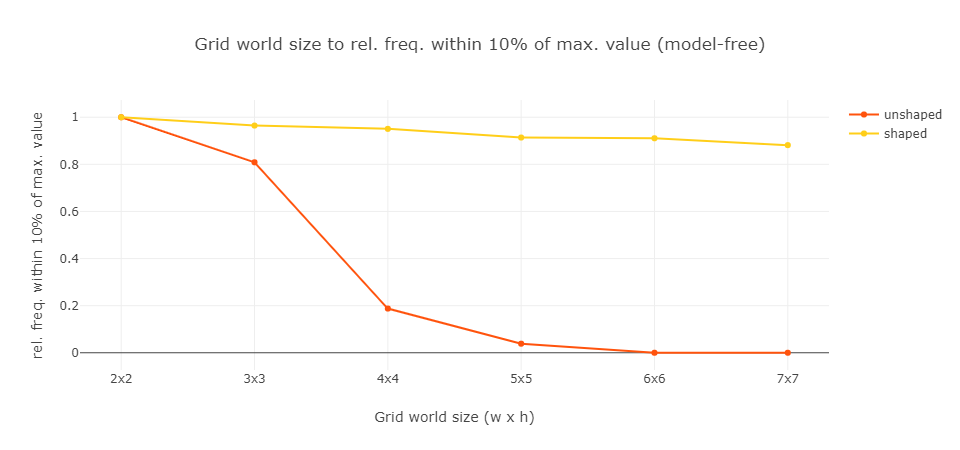}
        \caption{Rel. freq. within 10\% of max. value for unshaped and shaped rewards}
        \label{fig:exp-reward-shaping:top-ten-percent-increase}
    \end{figure}
\end{experiment}

%
%
\begin{experiment}{safety}{Safety}
    The goal of this experiment is to empirically measure the effects of safety properties on the performance of learning. Here, a preemptive shield \cite{alshiekh2017safe} is applied such that the agent is provided a list of safe actions upon action selection. Moreover, on-line compilation as outlined in \cref{alg:nmdp-to-mdp-on-line} is applied. Next, a $3 \times 3$ grid world is used and modeled as an RDP in which first-visit MC is applied as an RL learning algorithm. Transitions are deterministic to reduce experiment complexity. The quantitative measurement is defined by the performance of the learning algorithm. A total of 50 runs is performed, each of which consists of 1000 episodes and a maximum of 50 steps per episode. Furthermore, $\gamma = 1$ and $\epsilon = 0.1$. The agent always starts in state $s_{11}$ and an episode always terminates when the agent reaches state $s_{31}$. Goal $\langle\mathit{true}^{*};x_{is1} \land y_{is3};\mathit{true}^{*} \rangle\mathit{end}$ is specified for which the agent is rewarded $+50$ for reaching state $s_{13}$, i.e. the bottom-left state. A step cost of $-1$ is rewarded with each step the agent takes in the environment. An additional safety property $[\mathit{true}^{*}]\langle(\lnot x_{is1} \land \lnot y_{is2})^{*} \rangle\mathit{end}$ is specified in \LDLf{} where the agent should never visit unsafe state $s_{21}$. \Cref{fig:exp-safety} outlines possible optimal traces for unsafe and safe situations in \cref{fig:exp-safety:unsafe} and \cref{fig:exp-safety:safe} respectively. Note that terminal states, i.e. the top-right states in both \cref{fig:exp-safety:unsafe,fig:exp-safety:safe}, are marked solid black and that unsafe states, i.e. the center-left states in both \cref{fig:exp-safety:unsafe,fig:exp-safety:safe} are marked solid red.
    
    First, in order to verify no unsafe condition is met, \cref{fig:exp-safety:product-mdp} outlines the extended MDP for this experiment. Because of a technical index mapping, an unsafe state would have a label that starts with $(1,2,\ldots)$, corresponding to unsafe state $s_{12}$. As can be observed, there is no state $s^{\prime} \in S^{\prime}$ of extended MDP $M^{\prime}$ for which $\tau(s^{\prime}) \rightarrow s_{21}$. In other words, the unsafe state is never encountered. Therefore, safety property $[\mathit{true}^{*}]\langle(\lnot x_{is1} \land \lnot y_{is2})^{*} \rangle\mathit{end}$ is never violated. Furthermore, \Cref{fig:exp-safety:mf} plots the results for this experiment. Let us reconsider the results from \cref{fig:exp-safety-avg-return} and \cref{fig:exp-safety-ext-mdp-size}. It can be observed that, when learning performance is decreased, the size of the state space has become smaller. However, the intuition is that, when less states are to be explored, performance is generally increased. It appears, then, that RL performance is not necessarily dictated by the size of the state space. To account for what does impact the decreased learning performance, let us reconsider the experiment setup. Where state $s_{13}$ in \cref{fig:exp-safety:unsafe} can be reached from two states, i.e. $s_{12}$ and $s_{23}$, the same state can only be reached from a single state $s_{23}$ in \cref{fig:exp-safety:safe}. This observation leads to the conjecture that RL performance is related to the \emph{accessibility} of states required to satisfy goal formulae. That is, when states have less paths by which they can be reached, RL performance decreases.
    
    \begin{figure}[t]
        \centering
        \subfloat[avg. return]{
            \includegraphics[width=5.75cm]{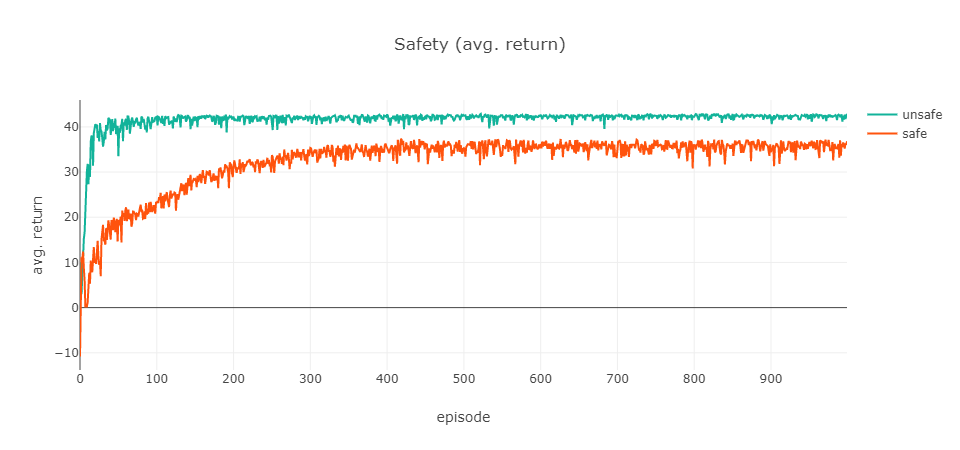}
            \label{fig:exp-safety-avg-return}
        }
        \subfloat[ext. MDP size]{
            \includegraphics[width=5.75cm]{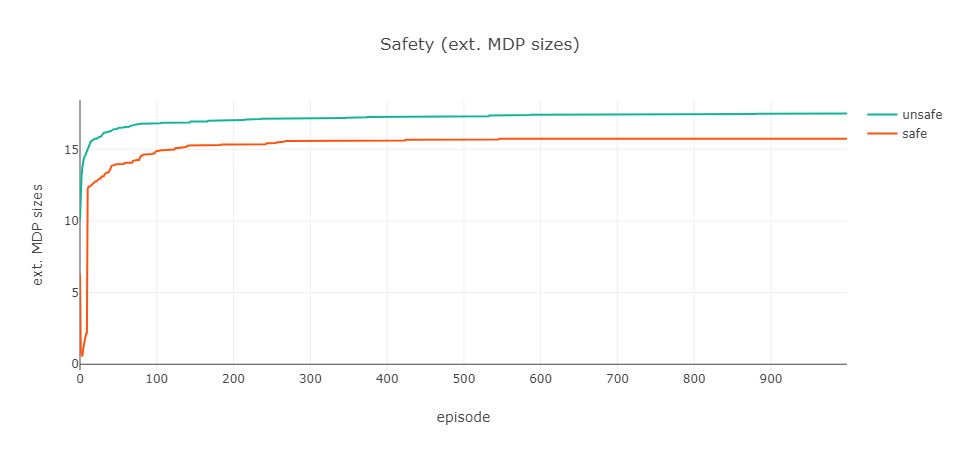}
            \label{fig:exp-safety-ext-mdp-size}
        }
        \hfill
        \caption{MC performance for safety}
        \label{fig:exp-safety:mf}
    \end{figure}
    
    \begin{figure}[t]
        \centering
        \includegraphics[width=9cm]{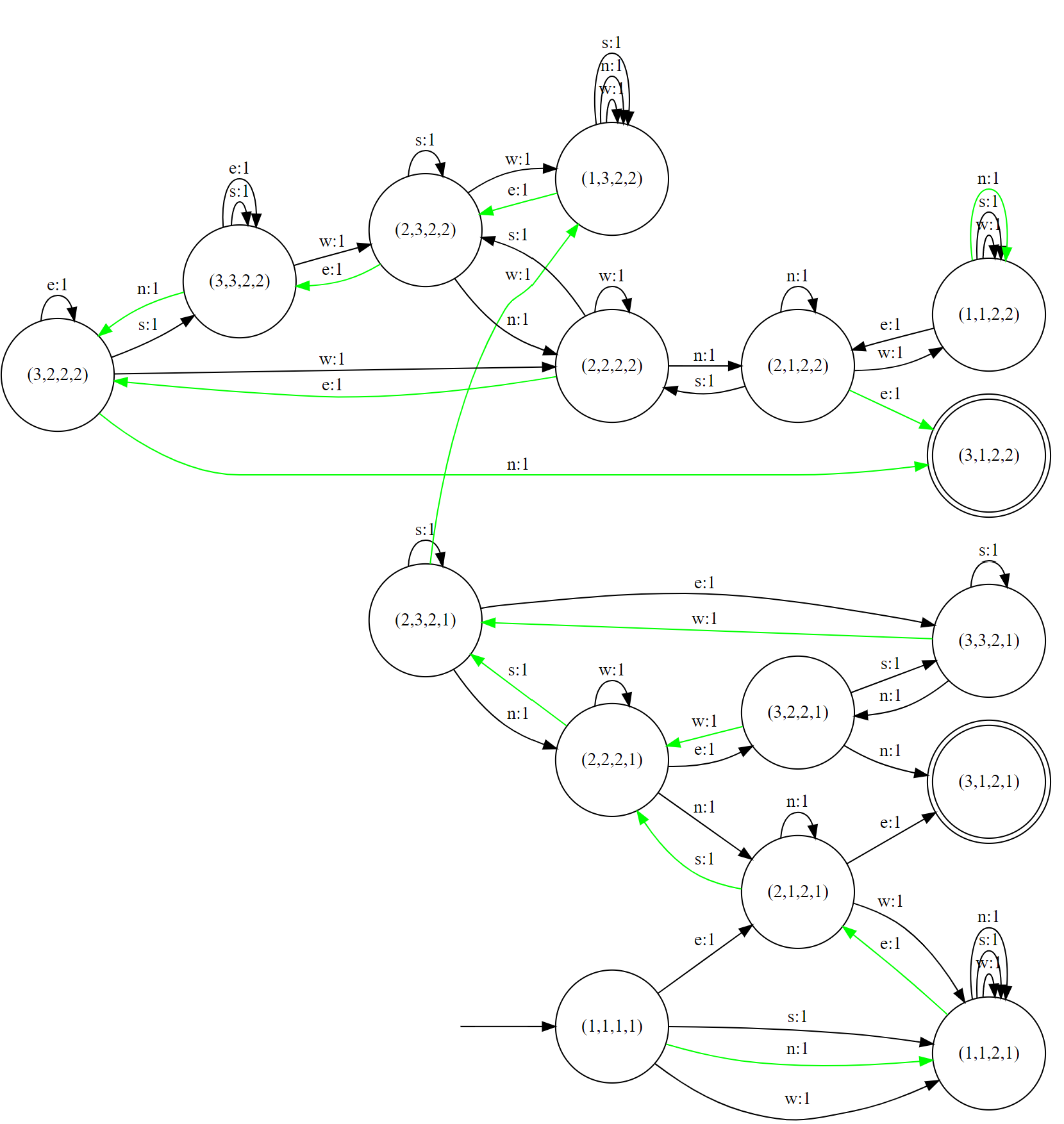}
        \caption{Product MDP for safety experiment}
        \label{fig:exp-safety:product-mdp}
    \end{figure}
\end{experiment}

%
%
\begin{experiment}{regular-transition}{Non-Markovian Transitions}
    This experiment focuses on non-Markovian transition models. Here, on-line compilation using \cref{alg:nmdp-to-mdp-on-line} will be used in a model-free setting, with first-visit MC. The quantitative measurements are defined by the averaged returns per episode, the averaged number of steps per episode and the size of the extended MDP. The experiment consists of 50 runs, each 1000 episodes with a maximum of 50 steps per episode. Furthermore, $\gamma = 1$ and $\epsilon = 0.1$. The grid world is defined by a $5 \times 2$ rectangle. The agent always starts an episode in state $s_{11}$ and state $s_{51}$ is defined as a terminal state. There is a single goal $\langle \mathit{true}^{*};x_{is3} \land y_{is1};\mathit{true}^{*} \rangle\mathit{end}$ rewarded when reaching $s_{41}$ of $+10$, with the addition of a step cost encoded in \LDLf{} valued as $-1$.
    
    \begin{figure}[t]
        \centering
        \subfloat[Non-deterministic]{
            \scalebox{.75}{\input{images/experiment-transition-complexity-non-det}}
            \label{fig:exp-transition-complexity:non-det}
        }
        \subfloat[Regular]{
            \scalebox{.75}{\input{images/experiment-transition-complexity-regular}}
            \label{fig:exp-transition-complexity:regular}
        }
        \hfill
        \caption{Possible optimal traces for transition complexity variations}
        \label{fig:exp-transition-complexity:setup}
    \end{figure}
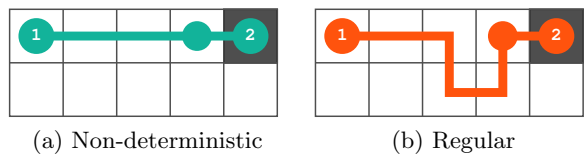
        
    The first set of 50 runs uses non-deterministic transitions. That is, in every state of the grid world the agent has a $0.8$ probability of ending up in the next state and a $0.2$ probability of remaining in its current state. For example, when in $s_{21}$ and taking action \textit{s}, there is a $0.8$ probability that we end up in $s_{22}$ and a $0.2$ probability to remain in $s_{21}$. The transition from $s_{31}$ for action \textit{e} is defined by $\langle \mathit{true}^{*};x_{is3} \land y_{is1} \rangle\mathit{end}$ and will be different for the regular transition defined below. A possible optimal trace is visually displayed in \cref{fig:exp-transition-complexity:non-det}.
    
    The second set of 50 runs uses regular transitions. Now, when the agent reaches $s_{31}$, a transition $\langle \mathit{true}^{*};x_{is2} \land y_{is1};x_{is3} \land y_{is1} \rangle\mathit{end}$ is defined that depends on whether the agent came from state $_{21}$ or not. In case the previous state was $s_{21}$, the transition of action \textit{e} in $s_{31}$ is defined as a $0.1$ probability of ending up in $s_{41}$ and a $0.9$ probability of remaining in $s_{31}$. Otherwise, the same transition probabilities used for the non-deterministic transitions apply, i.e. a $0.8$ probability of ending up in $s_{41}$ and a $0.2$ probability of remaining in $s_{31}$. Here, the transition in $s_{31}$ is regularly defined and depends on a history of states. A possible optimal trace for regular transitions is given in \cref{fig:exp-transition-complexity:regular}.
    
    \begin{figure}[t]
        \centering
        \subfloat[avg. return]{
            \includegraphics[width=5.75cm]{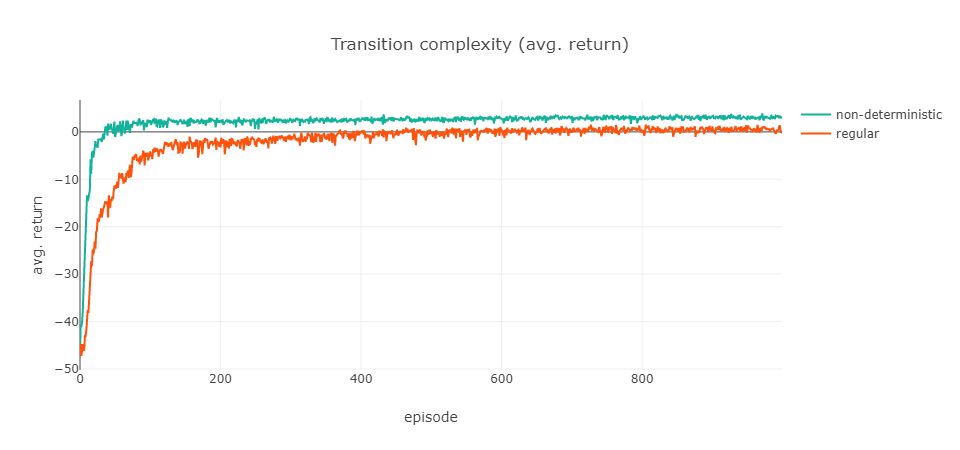}
            \label{fig:exp-transition-complexity-avg-return}
        }
        \subfloat[ext. MDP size]{
            \includegraphics[width=5.75cm]{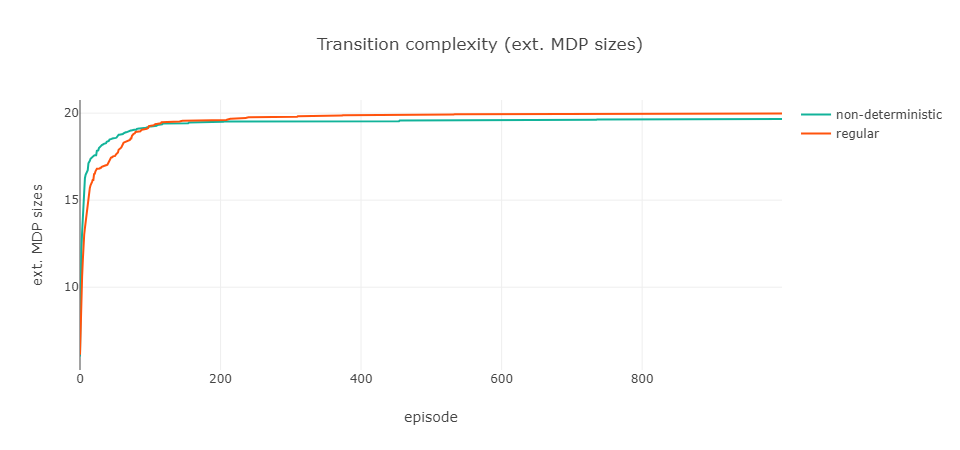}
            \label{fig:exp-transition-complexity-ext-mdp-size}
        }
        \hfill
        \caption{MC performance for transition complexity variations}
        \label{fig:exp-transition-complexity:mf}
    \end{figure}
    
    From \cref{fig:exp-transition-complexity:mf}, we observe that regular non-deterministic transitions, when compared to non-regular ones, induce a harder problem for a model-free setting, while the size of the state space is only increased by one additional state that keeps track of whether or not $s_{21}$ has been visited. In other words, a small increase in size can evidently generate a significantly harder problem.
\end{experiment}

%
%
\section{Conclusions and Future Work}\label{section:conclusions}
We have introduced a new tool chain to compute with regular decision processes, and experimented with novel algorithmic variations with the aim to gain insight in how complexity of temporal logic formulas relates to the complexity of learning algorithms such as MC RL for the resulting extended MDPs. We have shown that by increasing the world size for similar built formulas problems get harder (\textbf{R1})), but also that reward shaping on the automata representing those formulas can really help learning, \emph{and} exploration (\textbf{R2}). The safety experiments (\textbf{R3}) have shown less states do not necessarily result in easier learning tasks, and the non-Markov transitions experiments (\textbf{R4}) showed that these only caused a small increase in state space size, but did complicate learning a lot more.

Our overall conclusions of the experiments point to our main future work direction. It seems that there are complex relations between i) the complexity and properties of the temporal formulae defining the non-Markovian aspects, ii) the resulting size and connection structure of the extended MDP, and iii) the learning performance of online RL algorithms for the extended MDP. Much more work is needed to evaluate a temporal specification for a particular problem, and assess its influence on the complexity of learning the original task in the presence of the new rule. For MDPs there is much work on measures relating to e.g. homomorphism and abstraction \cite{van2009logic,wang2019measuring} and work is starting to emerge to gain more insight in the logical side \cite{romeo2018metric} but their interaction needs study.

Other future work should focus on \emph{representations} and \emph{applications}. For the first, there is much to be gained by utilizing existing formal methods, for example the use of transducers and Mealy machines \cite{KR2020-89} trading off the size of the state space with compositional modeling. Equally important is to focus more on utilizing model checking tools \cite{alshiekh2017safe,giaquinta2018strategy}. Application-wise, there are plenty of opportunities to utilize the methods in this paper, for example to constrain RL dialogue agents, in medical domains with logically represented medical guidance and regulations, or to implement coaching strategies in RL coaching agents \cite{el2018personalization}.

%
%
%
\bibliographystyle{splncs04}
\bibliography{paper}
\end{document}

%% file: images/grid-world-rdp-transition.tex
\tikzset{every picture/.style={line width=0.75pt}} 

\begin{tikzpicture}[x=0.75pt,y=0.75pt,yscale=-1,xscale=1]

\draw  [draw opacity=0][fill={rgb, 255:red, 78; green, 78; blue, 78 }  ,fill opacity=1 ] (250,20) -- (310,20) -- (310,80) -- (250,80) -- cycle ;
\draw  [draw opacity=0][line width=0.75]  (130,20) -- (310,20) -- (310,200) -- (130,200) -- cycle ; \draw  [color={rgb, 255:red, 78; green, 78; blue, 78 }  ,draw opacity=1 ][line width=0.75]  (190,20) -- (190,200)(250,20) -- (250,200) ; \draw  [color={rgb, 255:red, 78; green, 78; blue, 78 }  ,draw opacity=1 ][line width=0.75]  (130,80) -- (310,80)(130,140) -- (310,140) ; \draw  [color={rgb, 255:red, 78; green, 78; blue, 78 }  ,draw opacity=1 ][line width=0.75]  (130,20) -- (310,20) -- (310,200) -- (130,200) -- cycle ;
\draw [color={rgb, 255:red, 255; green, 83; blue, 13 }  ,draw opacity=1 ][line width=4.5]    (174,50) -- (174,110) -- (220,110) -- (220,156) -- (280,156) ;
\draw [shift={(280,156)}, rotate = 0] [color={rgb, 255:red, 255; green, 83; blue, 13 }  ,draw opacity=1 ][fill={rgb, 255:red, 255; green, 83; blue, 13 }  ,fill opacity=1 ][line width=4.5]      (0, 0) circle [x radius= 9.05, y radius= 9.05]   ;
\draw [shift={(174,50)}, rotate = 90] [color={rgb, 255:red, 255; green, 83; blue, 13 }  ,draw opacity=1 ][fill={rgb, 255:red, 255; green, 83; blue, 13 }  ,fill opacity=1 ][line width=4.5]      (0, 0) circle [x radius= 9.05, y radius= 9.05]   ;
\draw [color={rgb, 255:red, 18; green, 179; blue, 154 }  ,draw opacity=1 ][line width=4.5]    (146,50) -- (146,184) -- (280,184) ;
\draw [shift={(280,184)}, rotate = 0] [color={rgb, 255:red, 18; green, 179; blue, 154 }  ,draw opacity=1 ][fill={rgb, 255:red, 18; green, 179; blue, 154 }  ,fill opacity=1 ][line width=4.5]      (0, 0) circle [x radius= 9.05, y radius= 9.05]   ;
\draw [shift={(146,50)}, rotate = 90] [color={rgb, 255:red, 18; green, 179; blue, 154 }  ,draw opacity=1 ][fill={rgb, 255:red, 18; green, 179; blue, 154 }  ,fill opacity=1 ][line width=4.5]      (0, 0) circle [x radius= 9.05, y radius= 9.05]   ;
\draw [color={rgb, 255:red, 18; green, 179; blue, 154 }  ,draw opacity=1 ][line width=1.5]    (360,60) -- (380,60) ;
\draw [shift={(380,60)}, rotate = 0] [color={rgb, 255:red, 18; green, 179; blue, 154 }  ,draw opacity=1 ][fill={rgb, 255:red, 18; green, 179; blue, 154 }  ,fill opacity=1 ][line width=1.5]      (0, 0) circle [x radius= 4.36, y radius= 4.36]   ;
\draw [shift={(360,60)}, rotate = 0] [color={rgb, 255:red, 18; green, 179; blue, 154 }  ,draw opacity=1 ][fill={rgb, 255:red, 18; green, 179; blue, 154 }  ,fill opacity=1 ][line width=1.5]      (0, 0) circle [x radius= 4.36, y radius= 4.36]   ;
\draw [color={rgb, 255:red, 255; green, 83; blue, 13 }  ,draw opacity=1 ][line width=1.5]    (360,90) -- (380,90) ;
\draw [shift={(380,90)}, rotate = 0] [color={rgb, 255:red, 255; green, 83; blue, 13 }  ,draw opacity=1 ][fill={rgb, 255:red, 255; green, 83; blue, 13 }  ,fill opacity=1 ][line width=1.5]      (0, 0) circle [x radius= 4.36, y radius= 4.36]   ;
\draw [shift={(360,90)}, rotate = 0] [color={rgb, 255:red, 255; green, 83; blue, 13 }  ,draw opacity=1 ][fill={rgb, 255:red, 255; green, 83; blue, 13 }  ,fill opacity=1 ][line width=1.5]      (0, 0) circle [x radius= 4.36, y radius= 4.36]   ;

\draw (402,53) node [anchor=north west][inner sep=0.75pt]   [align=left] {$\displaystyle h_{1}$};
\draw (402,83) node [anchor=north west][inner sep=0.75pt]   [align=left] {$\displaystyle h_{2}$};
\draw (140,44) node [anchor=north west][inner sep=0.75pt]  [color={rgb, 255:red, 255; green, 255; blue, 255 }  ,opacity=1 ] [align=left] {\textbf{{\fontfamily{pcr}\selectfont 1}}};
\draw (169,44) node [anchor=north west][inner sep=0.75pt]  [color={rgb, 255:red, 255; green, 255; blue, 255 }  ,opacity=1 ] [align=left] {\textbf{{\fontfamily{pcr}\selectfont 1}}};
\draw (275,150) node [anchor=north west][inner sep=0.75pt]  [color={rgb, 255:red, 255; green, 255; blue, 255 }  ,opacity=1 ] [align=left] {\textbf{{\fontfamily{pcr}\selectfont 2}}};
\draw (275,178) node [anchor=north west][inner sep=0.75pt]  [color={rgb, 255:red, 255; green, 255; blue, 255 }  ,opacity=1 ] [align=left] {\textbf{{\fontfamily{pcr}\selectfont 2}}};

\end{tikzpicture}

%% file: images/project-overview.tex
\tikzset{every picture/.style={line width=0.75pt}} 

\begin{tikzpicture}[x=0.75pt,y=0.75pt,yscale=-1,xscale=1]

\draw  [color={rgb, 255:red, 78; green, 78; blue, 78 }  ,draw opacity=1 ] (0,50) -- (100,50) -- (100,360) -- (0,360) -- cycle ;
\draw  [color={rgb, 255:red, 78; green, 78; blue, 78 }  ,draw opacity=1 ] (160,50) -- (260,50) -- (260,210) -- (160,210) -- cycle ;
\draw  [color={rgb, 255:red, 78; green, 78; blue, 78 }  ,draw opacity=1 ] (180,90) -- (240,90) -- (240,130) -- (180,130) -- cycle ;
\draw  [color={rgb, 255:red, 78; green, 78; blue, 78 }  ,draw opacity=1 ] (180,150) -- (240,150) -- (240,190) -- (180,190) -- cycle ;
\draw  [color={rgb, 255:red, 78; green, 78; blue, 78 }  ,draw opacity=1 ] (320,50) -- (420,50) -- (420,210) -- (320,210) -- cycle ;
\draw  [color={rgb, 255:red, 78; green, 78; blue, 78 }  ,draw opacity=1 ] (160,260) -- (260,260) -- (260,360) -- (160,360) -- cycle ;
\draw [color={rgb, 255:red, 78; green, 78; blue, 78 }  ,draw opacity=1 ]   (100,130) -- (157,130) ;
\draw [shift={(160,130)}, rotate = 180] [fill={rgb, 255:red, 78; green, 78; blue, 78 }  ,fill opacity=1 ][line width=0.08]  [draw opacity=0] (8.93,-4.29) -- (0,0) -- (8.93,4.29) -- cycle    ;
\draw [color={rgb, 255:red, 78; green, 78; blue, 78 }  ,draw opacity=1 ]   (260,130) -- (317,130) ;
\draw [shift={(320,130)}, rotate = 180] [fill={rgb, 255:red, 78; green, 78; blue, 78 }  ,fill opacity=1 ][line width=0.08]  [draw opacity=0] (8.93,-4.29) -- (0,0) -- (8.93,4.29) -- cycle    ;
\draw [color={rgb, 255:red, 78; green, 78; blue, 78 }  ,draw opacity=1 ]   (420,130) -- (467,130) ;
\draw [shift={(470,130)}, rotate = 180] [fill={rgb, 255:red, 78; green, 78; blue, 78 }  ,fill opacity=1 ][line width=0.08]  [draw opacity=0] (8.93,-4.29) -- (0,0) -- (8.93,4.29) -- cycle    ;
\draw [color={rgb, 255:red, 78; green, 78; blue, 78 }  ,draw opacity=1 ]   (100,310) -- (157,310) ;
\draw [shift={(160,310)}, rotate = 180] [fill={rgb, 255:red, 78; green, 78; blue, 78 }  ,fill opacity=1 ][line width=0.08]  [draw opacity=0] (8.93,-4.29) -- (0,0) -- (8.93,4.29) -- cycle    ;
\draw [color={rgb, 255:red, 78; green, 78; blue, 78 }  ,draw opacity=1 ]   (260,310) -- (317,310) ;
\draw [shift={(320,310)}, rotate = 180] [fill={rgb, 255:red, 78; green, 78; blue, 78 }  ,fill opacity=1 ][line width=0.08]  [draw opacity=0] (8.93,-4.29) -- (0,0) -- (8.93,4.29) -- cycle    ;
\draw [color={rgb, 255:red, 78; green, 78; blue, 78 }  ,draw opacity=1 ]   (420,310) -- (557,310) ;
\draw [shift={(560,310)}, rotate = 180] [fill={rgb, 255:red, 78; green, 78; blue, 78 }  ,fill opacity=1 ][line width=0.08]  [draw opacity=0] (8.93,-4.29) -- (0,0) -- (8.93,4.29) -- cycle    ;
\draw [color={rgb, 255:red, 78; green, 78; blue, 78 }  ,draw opacity=1 ]   (520,180) -- (520,310) ;
\draw [color={rgb, 255:red, 78; green, 78; blue, 78 }  ,draw opacity=1 ]   (370,210) -- (370,257) ;
\draw [shift={(370,260)}, rotate = 270] [fill={rgb, 255:red, 78; green, 78; blue, 78 }  ,fill opacity=1 ][line width=0.08]  [draw opacity=0] (8.93,-4.29) -- (0,0) -- (8.93,4.29) -- cycle    ;
\draw  [color={rgb, 255:red, 78; green, 78; blue, 78 }  ,draw opacity=1 ] (370,260) -- (420,310) -- (370,360) -- (320,310) -- cycle ;
\draw  [color={rgb, 255:red, 78; green, 78; blue, 78 }  ,draw opacity=1 ] (520,80) -- (570,130) -- (520,180) -- (470,130) -- cycle ;
\draw  [color={rgb, 255:red, 78; green, 78; blue, 78 }  ,draw opacity=1 ] (560,310) .. controls (560,282.39) and (582.39,260) .. (610,260) .. controls (637.61,260) and (660,282.39) .. (660,310) .. controls (660,337.61) and (637.61,360) .. (610,360) .. controls (582.39,360) and (560,337.61) .. (560,310) -- cycle ;
\draw  [color={rgb, 255:red, 78; green, 78; blue, 78 }  ,draw opacity=1 ] (20,90) -- (80,90) -- (80,130) -- (20,130) -- cycle ;
\draw  [color={rgb, 255:red, 78; green, 78; blue, 78 }  ,draw opacity=1 ] (20,150) -- (80,150) -- (80,190) -- (20,190) -- cycle ;
\draw [color={rgb, 255:red, 78; green, 78; blue, 78 }  ,draw opacity=1 ]   (210,20) -- (210,47) ;
\draw [shift={(210,50)}, rotate = 270] [fill={rgb, 255:red, 78; green, 78; blue, 78 }  ,fill opacity=1 ][line width=0.08]  [draw opacity=0] (8.93,-4.29) -- (0,0) -- (8.93,4.29) -- cycle    ;
\draw [color={rgb, 255:red, 78; green, 78; blue, 78 }  ,draw opacity=1 ]   (570,130) -- (610,130) -- (610,257) ;
\draw [shift={(610,260)}, rotate = 270] [fill={rgb, 255:red, 78; green, 78; blue, 78 }  ,fill opacity=1 ][line width=0.08]  [draw opacity=0] (8.93,-4.29) -- (0,0) -- (8.93,4.29) -- cycle    ;

\draw (11,62) node [anchor=north west][inner sep=0.75pt]  [font=\footnotesize,color={rgb, 255:red, 78; green, 78; blue, 78 }  ,opacity=1 ] [align=left] {Temp. Logic};
\draw (171,63) node [anchor=north west][inner sep=0.75pt]  [font=\footnotesize,color={rgb, 255:red, 78; green, 78; blue, 78 }  ,opacity=1 ] [align=left] {NMDP};
\draw (185,102) node [anchor=north west][inner sep=0.75pt]  [font=\footnotesize,color={rgb, 255:red, 78; green, 78; blue, 78 }  ,opacity=1 ] [align=left] {NMRDP};
\draw (185,162) node [anchor=north west][inner sep=0.75pt]  [font=\footnotesize,color={rgb, 255:red, 78; green, 78; blue, 78 }  ,opacity=1 ] [align=left] {RDP};
\draw (331,62) node [anchor=north west][inner sep=0.75pt]  [font=\footnotesize,color={rgb, 255:red, 78; green, 78; blue, 78 }  ,opacity=1 ] [align=left] {MDP};
\draw (496,122) node [anchor=north west][inner sep=0.75pt]  [font=\footnotesize,color={rgb, 255:red, 78; green, 78; blue, 78 }  ,opacity=1 ] [align=left] {RL / DP};
\draw (171,272) node [anchor=north west][inner sep=0.75pt]  [font=\footnotesize,color={rgb, 255:red, 78; green, 78; blue, 78 }  ,opacity=1 ] [align=left] {Properties};
\draw (341,292) node [anchor=north west][inner sep=0.75pt]  [font=\footnotesize,color={rgb, 255:red, 78; green, 78; blue, 78 }  ,opacity=1 ] [align=left] {\begin{minipage}[lt]{36.74pt}\setlength\topsep{0pt}
\begin{center}
Model\\Checking
\end{center}

\end{minipage}};
\draw (591,302) node [anchor=north west][inner sep=0.75pt]  [font=\footnotesize,color={rgb, 255:red, 78; green, 78; blue, 78 }  ,opacity=1 ] [align=left] {Policy};
\draw (114,112) node [anchor=north west][inner sep=0.75pt]  [font=\scriptsize,color={rgb, 255:red, 78; green, 78; blue, 78 }  ,opacity=1 ] [align=left] {input};
\draw (271,112) node [anchor=north west][inner sep=0.75pt]  [font=\scriptsize,color={rgb, 255:red, 78; green, 78; blue, 78 }  ,opacity=1 ] [align=left] {compile};
\draw (434,112) node [anchor=north west][inner sep=0.75pt]  [font=\scriptsize,color={rgb, 255:red, 78; green, 78; blue, 78 }  ,opacity=1 ] [align=left] {input};
\draw (114,292) node [anchor=north west][inner sep=0.75pt]  [font=\scriptsize,color={rgb, 255:red, 78; green, 78; blue, 78 }  ,opacity=1 ] [align=left] {define};
\draw (274,292) node [anchor=north west][inner sep=0.75pt]  [font=\scriptsize,color={rgb, 255:red, 78; green, 78; blue, 78 }  ,opacity=1 ] [align=left] {input};
\draw (434,292) node [anchor=north west][inner sep=0.75pt]  [font=\scriptsize,color={rgb, 255:red, 78; green, 78; blue, 78 }  ,opacity=1 ] [align=left] {output};
\draw (528,217) node [anchor=north west][inner sep=0.75pt]  [font=\scriptsize,color={rgb, 255:red, 78; green, 78; blue, 78 }  ,opacity=1 ] [align=left] {output};
\draw (374,227) node [anchor=north west][inner sep=0.75pt]  [font=\scriptsize,color={rgb, 255:red, 78; green, 78; blue, 78 }  ,opacity=1 ] [align=left] {input};
\draw (31,102) node [anchor=north west][inner sep=0.75pt]  [font=\footnotesize,color={rgb, 255:red, 78; green, 78; blue, 78 }  ,opacity=1 ] [align=left] {LTL$\displaystyle _{f}$};
\draw (31,162) node [anchor=north west][inner sep=0.75pt]  [font=\footnotesize,color={rgb, 255:red, 78; green, 78; blue, 78 }  ,opacity=1 ] [align=left] {LDL$\displaystyle _{f}$};
\draw (219,27) node [anchor=north west][inner sep=0.75pt]  [font=\scriptsize,color={rgb, 255:red, 78; green, 78; blue, 78 }  ,opacity=1 ] [align=left] {input};
\draw (571,112) node [anchor=north west][inner sep=0.75pt]  [font=\scriptsize,color={rgb, 255:red, 78; green, 78; blue, 78 }  ,opacity=1 ] [align=left] {output};

\end{tikzpicture}

%% file: algorithms/nmdp-to-mdp-off-line.tex
\SetKwInOut{Input}{input}
\SetKwInOut{Output}{output}
\Input{NMDP $M = \langle S, A, T, R \rangle$ with \LDLf{} reward automata $Q^{R}_{i}$ and \LDLf{} transition automata $Q^{T}_{j}$, with $Q_{k} \leftarrow Q^{R}_{i} \cup Q^{T}_{j}$ for convenience}
\Output{Extended MDP $M^{\prime} = \langle S^{\prime}, A^{\prime}, T^{\prime}, R^{\prime} \rangle$}
$t \leftarrow 0$; $s^{\prime}_{t} \leftarrow s^{\prime}_{0} \leftarrow (q_{1,0}, q_{2,0}, \ldots, q_{k,0}, s_{0})$; $A^{\prime} \leftarrow A$; $S, T, R \leftarrow \emptyset$\\
\While{$s^{\prime}_{t} \notin S^{\prime}$}{
    $s_{t} \leftarrow \tau(s^{\prime}_{t})$\\
    \For{$a \in A(s_{t})$}{
        $s_{t+1} \leftarrow T(\Lb(s_{t}), a)$\\
        \For{$q_{k, t} \in Q_{k, t}$}{
            $q_{k, t + 1} \leftarrow \mathtt{transition}(q_{k, t}, \Lb(s_{t + 1}))$
        }
        $s^{\prime}_{t+1} \leftarrow (q_{1,t + 1}, q_{2,t + 1}, \ldots, q_{k,t + 1}, s_{t + 1})$\\
        $S^{\prime} \leftarrow S^{\prime} \cup \{s^{\prime}_{t + 1}\}$\\
        $T^{\prime}(s^{\prime}_{t}, a, s^{\prime}_{t + 1}) \leftarrow T(\Lb(s_{t}), a, \Lb(s_{t + 1}))$\label{line:nmdp-to-mdp-off-line:transition}\\
        $R^{\prime}(s^{\prime}_{t}, a, s^{\prime}_{t + 1}) \leftarrow \mathtt{sum\_accept}(Q^{R}_{i, t + 1})$\\
        $s^{\prime}_{t} \leftarrow s^{\prime}_{t + 1}$\\
    }
}
\Return $M^{\prime}$

%% file: algorithms/nmdp-to-mdp-on-line.tex
\SetKwInOut{Input}{input}
\SetKwInOut{Output}{output}
\SetKwRepeat{Repeat}{repeat}{until}

\Input{Environment $\mathtt{env}$ with $n$-step limit per episode, exploration factor $\epsilon$, discount factor $\gamma$ and $\mathtt{max\_episodes}$}
\Output{Policy $\pi$}
$s_{t} \leftarrow s_{0} \leftarrow \mathtt{env.reset()}$; $Q(s, a) \leftarrow \mathtt{arbitrary()}$ for all $s \in S, a \in A(s)$; $\pi \leftarrow \mathtt{arbitrary()}$\\
\Repeat{$\mathtt{max\_episodes}$}{
    $\mathtt{ep} = \mathtt{generate\_episode}(n, A(s_{t}), \pi, \epsilon)$\\
    
    $T \leftarrow |\mathtt{ep}|$\\
    $G \leftarrow 0$\\
    \ForEach{step of $\mathtt{ep}$, $t = T - 1, T - 2, \ldots, 0$}{
        $G \leftarrow \gamma G + r_{t + 1}$\\
        $Q(s_{t}, a_{t}) \leftarrow G$\\
        $\pi(s_{t}) \leftarrow \underset{a}{\text{arg max}}(Q(s_{t}, a))$\\
    }
}
\Return $\pi$

%% file: images/experiment-goal-sparsity-adjacent.tex
\tikzset{every picture/.style={line width=0.75pt}} 

\begin{tikzpicture}[x=0.75pt,y=0.75pt,yscale=-1,xscale=1]

\draw  [fill={rgb, 255:red, 78; green, 78; blue, 78 }  ,fill opacity=1 ] (262,80) -- (298,80) -- (298,116.05) -- (262,116.05) -- cycle ;
\draw  [draw opacity=0][line width=0.75]  (190,80) -- (298,80) -- (298,188) -- (190,188) -- cycle ; \draw  [color={rgb, 255:red, 78; green, 78; blue, 78 }  ,draw opacity=1 ][line width=0.75]  (226,80) -- (226,188)(262,80) -- (262,188) ; \draw  [color={rgb, 255:red, 78; green, 78; blue, 78 }  ,draw opacity=1 ][line width=0.75]  (190,116) -- (298,116)(190,152) -- (298,152) ; \draw  [color={rgb, 255:red, 78; green, 78; blue, 78 }  ,draw opacity=1 ][line width=0.75]  (190,80) -- (298,80) -- (298,188) -- (190,188) -- cycle ;
\draw [color={rgb, 255:red, 18; green, 179; blue, 154 }  ,draw opacity=1 ][line width=4.5]    (208,98) -- (208,135) -- (244,135) -- (244,170) -- (280,170) -- (280,98) ;
\draw [shift={(280,98)}, rotate = 270] [color={rgb, 255:red, 18; green, 179; blue, 154 }  ,draw opacity=1 ][fill={rgb, 255:red, 18; green, 179; blue, 154 }  ,fill opacity=1 ][line width=4.5]      (0, 0) circle [x radius= 9.05, y radius= 9.05]   ;
\draw [shift={(208,98)}, rotate = 90] [color={rgb, 255:red, 18; green, 179; blue, 154 }  ,draw opacity=1 ][fill={rgb, 255:red, 18; green, 179; blue, 154 }  ,fill opacity=1 ][line width=4.5]      (0, 0) circle [x radius= 9.05, y radius= 9.05]   ;
\draw  [draw opacity=0][fill={rgb, 255:red, 18; green, 179; blue, 154 }  ,fill opacity=1 ] (234,170) .. controls (234,164.48) and (238.48,160) .. (244,160) .. controls (249.52,160) and (254,164.48) .. (254,170) .. controls (254,175.52) and (249.52,180) .. (244,180) .. controls (238.48,180) and (234,175.52) .. (234,170) -- cycle ;
\draw  [draw opacity=0][fill={rgb, 255:red, 18; green, 179; blue, 154 }  ,fill opacity=1 ] (270,170) .. controls (270,164.48) and (274.48,160) .. (280,160) .. controls (285.52,160) and (290,164.48) .. (290,170) .. controls (290,175.52) and (285.52,180) .. (280,180) .. controls (274.48,180) and (270,175.52) .. (270,170) -- cycle ;

\draw (203,92) node [anchor=north west][inner sep=0.75pt]  [color={rgb, 255:red, 255; green, 255; blue, 255 }  ,opacity=1 ] [align=left] {\textbf{{\fontfamily{pcr}\selectfont 1}}};
\draw (275,92) node [anchor=north west][inner sep=0.75pt]  [color={rgb, 255:red, 255; green, 255; blue, 255 }  ,opacity=1 ] [align=left] {\textbf{{\fontfamily{pcr}\selectfont 2}}};

\end{tikzpicture}

%% file: images/experiment-goal-sparsity-center.tex
\tikzset{every picture/.style={line width=0.75pt}} 

\begin{tikzpicture}[x=0.75pt,y=0.75pt,yscale=-1,xscale=1]

\draw  [fill={rgb, 255:red, 78; green, 78; blue, 78 }  ,fill opacity=1 ] (262,80) -- (298,80) -- (298,116.05) -- (262,116.05) -- cycle ;
\draw  [draw opacity=0][line width=0.75]  (190,80) -- (298,80) -- (298,188) -- (190,188) -- cycle ; \draw  [color={rgb, 255:red, 78; green, 78; blue, 78 }  ,draw opacity=1 ][line width=0.75]  (226,80) -- (226,188)(262,80) -- (262,188) ; \draw  [color={rgb, 255:red, 78; green, 78; blue, 78 }  ,draw opacity=1 ][line width=0.75]  (190,116) -- (298,116)(190,152) -- (298,152) ; \draw  [color={rgb, 255:red, 78; green, 78; blue, 78 }  ,draw opacity=1 ][line width=0.75]  (190,80) -- (298,80) -- (298,188) -- (190,188) -- cycle ;
\draw [color={rgb, 255:red, 255; green, 83; blue, 13 }  ,draw opacity=1 ][line width=4.5]    (208,98) -- (208,170) -- (240,170) -- (240,130) -- (250,130) -- (250,170) -- (280,170) -- (280,98) ;
\draw [shift={(280,98)}, rotate = 270] [color={rgb, 255:red, 255; green, 83; blue, 13 }  ,draw opacity=1 ][fill={rgb, 255:red, 255; green, 83; blue, 13 }  ,fill opacity=1 ][line width=4.5]      (0, 0) circle [x radius= 9.05, y radius= 9.05]   ;
\draw [shift={(208,98)}, rotate = 90] [color={rgb, 255:red, 255; green, 83; blue, 13 }  ,draw opacity=1 ][fill={rgb, 255:red, 255; green, 83; blue, 13 }  ,fill opacity=1 ][line width=4.5]      (0, 0) circle [x radius= 9.05, y radius= 9.05]   ;
\draw  [draw opacity=0][fill={rgb, 255:red, 255; green, 83; blue, 13 }  ,fill opacity=1 ] (198,170) .. controls (198,164.48) and (202.48,160) .. (208,160) .. controls (213.52,160) and (218,164.48) .. (218,170) .. controls (218,175.52) and (213.52,180) .. (208,180) .. controls (202.48,180) and (198,175.52) .. (198,170) -- cycle ;
\draw  [draw opacity=0][fill={rgb, 255:red, 255; green, 83; blue, 13 }  ,fill opacity=1 ] (235,130) .. controls (235,124.48) and (239.48,120) .. (245,120) .. controls (250.52,120) and (255,124.48) .. (255,130) .. controls (255,135.52) and (250.52,140) .. (245,140) .. controls (239.48,140) and (235,135.52) .. (235,130) -- cycle ;
\draw  [draw opacity=0][fill={rgb, 255:red, 255; green, 83; blue, 13 }  ,fill opacity=1 ] (270,170) .. controls (270,164.48) and (274.48,160) .. (280,160) .. controls (285.52,160) and (290,164.48) .. (290,170) .. controls (290,175.52) and (285.52,180) .. (280,180) .. controls (274.48,180) and (270,175.52) .. (270,170) -- cycle ;

\draw (203,92) node [anchor=north west][inner sep=0.75pt]  [color={rgb, 255:red, 255; green, 255; blue, 255 }  ,opacity=1 ] [align=left] {\textbf{{\fontfamily{pcr}\selectfont 1}}};
\draw (275,92) node [anchor=north west][inner sep=0.75pt]  [color={rgb, 255:red, 255; green, 255; blue, 255 }  ,opacity=1 ] [align=left] {\textbf{{\fontfamily{pcr}\selectfont 2}}};

\end{tikzpicture}

%% file: images/experiment-goal-sparsity-diagonal.tex
\tikzset{every picture/.style={line width=0.75pt}} 

\begin{tikzpicture}[x=0.75pt,y=0.75pt,yscale=-1,xscale=1]

\draw  [fill={rgb, 255:red, 78; green, 78; blue, 78 }  ,fill opacity=1 ] (262,80) -- (298,80) -- (298,116.05) -- (262,116.05) -- cycle ;
\draw  [draw opacity=0][line width=0.75]  (190,80) -- (298,80) -- (298,188) -- (190,188) -- cycle ; \draw  [color={rgb, 255:red, 78; green, 78; blue, 78 }  ,draw opacity=1 ][line width=0.75]  (226,80) -- (226,188)(262,80) -- (262,188) ; \draw  [color={rgb, 255:red, 78; green, 78; blue, 78 }  ,draw opacity=1 ][line width=0.75]  (190,116) -- (298,116)(190,152) -- (298,152) ; \draw  [color={rgb, 255:red, 78; green, 78; blue, 78 }  ,draw opacity=1 ][line width=0.75]  (190,80) -- (298,80) -- (298,188) -- (190,188) -- cycle ;
\draw [color={rgb, 255:red, 255; green, 206; blue, 25 }  ,draw opacity=1 ][line width=4.5]    (208,98) -- (208,140) -- (240,140) -- (240,176) -- (280,176) -- (280,164) -- (250,164) -- (250,130) -- (220,130) -- (220,110) -- (280,110) -- (280,98) ;
\draw [shift={(280,98)}, rotate = 270] [color={rgb, 255:red, 255; green, 206; blue, 25 }  ,draw opacity=1 ][fill={rgb, 255:red, 255; green, 206; blue, 25 }  ,fill opacity=1 ][line width=4.5]      (0, 0) circle [x radius= 9.05, y radius= 9.05]   ;
\draw [shift={(208,98)}, rotate = 90] [color={rgb, 255:red, 255; green, 206; blue, 25 }  ,draw opacity=1 ][fill={rgb, 255:red, 255; green, 206; blue, 25 }  ,fill opacity=1 ][line width=4.5]      (0, 0) circle [x radius= 9.05, y radius= 9.05]   ;
\draw  [draw opacity=0][fill={rgb, 255:red, 255; green, 206; blue, 25 }  ,fill opacity=1 ] (270,170) .. controls (270,164.48) and (274.48,160) .. (280,160) .. controls (285.52,160) and (290,164.48) .. (290,170) .. controls (290,175.52) and (285.52,180) .. (280,180) .. controls (274.48,180) and (270,175.52) .. (270,170) -- cycle ;

\draw (203,92) node [anchor=north west][inner sep=0.75pt]  [color={rgb, 255:red, 255; green, 255; blue, 255 }  ,opacity=1 ] [align=left] {\textbf{{\fontfamily{pcr}\selectfont 1}}};
\draw (275,92) node [anchor=north west][inner sep=0.75pt]  [color={rgb, 255:red, 255; green, 255; blue, 255 }  ,opacity=1 ] [align=left] {\textbf{{\fontfamily{pcr}\selectfont 2}}};

\end{tikzpicture}

%% file: images/experiment-reward-shaping.tex
\tikzset{every picture/.style={line width=0.75pt}} 

\begin{tikzpicture}[x=0.75pt,y=0.75pt,yscale=-1,xscale=1]

\draw  [fill={rgb, 255:red, 78; green, 78; blue, 78 }  ,fill opacity=1 ] (334,80) -- (370,80) -- (370,116.05) -- (334,116.05) -- cycle ;
\draw  [draw opacity=0][line width=0.75]  (190,80) -- (370,80) -- (370,260) -- (190,260) -- cycle ; \draw  [color={rgb, 255:red, 78; green, 78; blue, 78 }  ,draw opacity=1 ][line width=0.75]  (226,80) -- (226,260)(262,80) -- (262,260)(298,80) -- (298,260)(334,80) -- (334,260) ; \draw  [color={rgb, 255:red, 78; green, 78; blue, 78 }  ,draw opacity=1 ][line width=0.75]  (190,116) -- (370,116)(190,152) -- (370,152)(190,188) -- (370,188)(190,224) -- (370,224) ; \draw  [color={rgb, 255:red, 78; green, 78; blue, 78 }  ,draw opacity=1 ][line width=0.75]  (190,80) -- (370,80) -- (370,260) -- (190,260) -- cycle ;
\draw [color={rgb, 255:red, 18; green, 179; blue, 154 }  ,draw opacity=1 ][line width=4.5]    (208,98) -- (208,242) -- (270,242) -- (270,170) -- (290,170) -- (290,242) -- (352,242) -- (352,98) ;
\draw [shift={(352,98)}, rotate = 270] [color={rgb, 255:red, 18; green, 179; blue, 154 }  ,draw opacity=1 ][fill={rgb, 255:red, 18; green, 179; blue, 154 }  ,fill opacity=1 ][line width=4.5]      (0, 0) circle [x radius= 9.05, y radius= 9.05]   ;
\draw [shift={(208,98)}, rotate = 90] [color={rgb, 255:red, 18; green, 179; blue, 154 }  ,draw opacity=1 ][fill={rgb, 255:red, 18; green, 179; blue, 154 }  ,fill opacity=1 ][line width=4.5]      (0, 0) circle [x radius= 9.05, y radius= 9.05]   ;
\draw  [draw opacity=0][fill={rgb, 255:red, 18; green, 179; blue, 154 }  ,fill opacity=1 ] (198,242) .. controls (198,236.48) and (202.48,232) .. (208,232) .. controls (213.52,232) and (218,236.48) .. (218,242) .. controls (218,247.52) and (213.52,252) .. (208,252) .. controls (202.48,252) and (198,247.52) .. (198,242) -- cycle ;
\draw  [draw opacity=0][fill={rgb, 255:red, 18; green, 179; blue, 154 }  ,fill opacity=1 ] (270,170) .. controls (270,164.48) and (274.48,160) .. (280,160) .. controls (285.52,160) and (290,164.48) .. (290,170) .. controls (290,175.52) and (285.52,180) .. (280,180) .. controls (274.48,180) and (270,175.52) .. (270,170) -- cycle ;
\draw  [draw opacity=0][fill={rgb, 255:red, 18; green, 179; blue, 154 }  ,fill opacity=1 ] (342,242) .. controls (342,236.48) and (346.48,232) .. (352,232) .. controls (357.52,232) and (362,236.48) .. (362,242) .. controls (362,247.52) and (357.52,252) .. (352,252) .. controls (346.48,252) and (342,247.52) .. (342,242) -- cycle ;

\draw (203,92) node [anchor=north west][inner sep=0.75pt]  [color={rgb, 255:red, 255; green, 255; blue, 255 }  ,opacity=1 ] [align=left] {\textbf{{\fontfamily{pcr}\selectfont 1}}};
\draw (347,92) node [anchor=north west][inner sep=0.75pt]  [color={rgb, 255:red, 255; green, 255; blue, 255 }  ,opacity=1 ] [align=left] {\textbf{{\fontfamily{pcr}\selectfont 2}}};

\end{tikzpicture}

%% file: images/experiment-safety-unsafe.tex
\tikzset{every picture/.style={line width=0.75pt}} 

\begin{tikzpicture}[x=0.75pt,y=0.75pt,yscale=-1,xscale=1]

\draw  [color={rgb, 255:red, 0; green, 0; blue, 0 }  ,draw opacity=1 ][fill={rgb, 255:red, 255; green, 0; blue, 0 }  ,fill opacity=1 ] (170,96) -- (206,96) -- (206,132.05) -- (170,132.05) -- cycle ;
\draw  [fill={rgb, 255:red, 78; green, 78; blue, 78 }  ,fill opacity=1 ] (242,60) -- (278,60) -- (278,96.05) -- (242,96.05) -- cycle ;
\draw  [draw opacity=0][line width=0.75]  (170,60) -- (278,60) -- (278,168) -- (170,168) -- cycle ; \draw  [color={rgb, 255:red, 78; green, 78; blue, 78 }  ,draw opacity=1 ][line width=0.75]  (206,60) -- (206,168)(242,60) -- (242,168) ; \draw  [color={rgb, 255:red, 78; green, 78; blue, 78 }  ,draw opacity=1 ][line width=0.75]  (170,96) -- (278,96)(170,132) -- (278,132) ; \draw  [color={rgb, 255:red, 78; green, 78; blue, 78 }  ,draw opacity=1 ][line width=0.75]  (170,60) -- (278,60) -- (278,168) -- (170,168) -- cycle ;
\draw [color={rgb, 255:red, 18; green, 179; blue, 154 }  ,draw opacity=1 ][line width=4.5]    (188.02,79.05) -- (189,140) -- (189,160) -- (230,160) -- (230,79) -- (260,79.02) ;
\draw [shift={(260,79.02)}, rotate = 0.05] [color={rgb, 255:red, 18; green, 179; blue, 154 }  ,draw opacity=1 ][fill={rgb, 255:red, 18; green, 179; blue, 154 }  ,fill opacity=1 ][line width=4.5]      (0, 0) circle [x radius= 9.05, y radius= 9.05]   ;
\draw [shift={(188.02,79.05)}, rotate = 89.08] [color={rgb, 255:red, 18; green, 179; blue, 154 }  ,draw opacity=1 ][fill={rgb, 255:red, 18; green, 179; blue, 154 }  ,fill opacity=1 ][line width=4.5]      (0, 0) circle [x radius= 9.05, y radius= 9.05]   ;
\draw  [draw opacity=0][fill={rgb, 255:red, 18; green, 179; blue, 154 }  ,fill opacity=1 ] (178,150) .. controls (178,144.48) and (182.48,140) .. (188,140) .. controls (193.52,140) and (198,144.48) .. (198,150) .. controls (198,155.52) and (193.52,160) .. (188,160) .. controls (182.48,160) and (178,155.52) .. (178,150) -- cycle ;

\draw (183,73) node [anchor=north west][inner sep=0.75pt]  [color={rgb, 255:red, 255; green, 255; blue, 255 }  ,opacity=1 ] [align=left] {\textbf{{\fontfamily{pcr}\selectfont 1}}};
\draw (255,73) node [anchor=north west][inner sep=0.75pt]  [color={rgb, 255:red, 255; green, 255; blue, 255 }  ,opacity=1 ] [align=left] {\textbf{{\fontfamily{pcr}\selectfont 2}}};

\end{tikzpicture}

%% file: images/experiment-safety-safe.tex
\tikzset{every picture/.style={line width=0.75pt}} 

\begin{tikzpicture}[x=0.75pt,y=0.75pt,yscale=-1,xscale=1]

\draw  [fill={rgb, 255:red, 78; green, 78; blue, 78 }  ,fill opacity=1 ] (242,60) -- (278,60) -- (278,96.05) -- (242,96.05) -- cycle ;
\draw  [draw opacity=0][line width=0.75]  (170,60) -- (278,60) -- (278,168) -- (170,168) -- cycle ; \draw  [color={rgb, 255:red, 78; green, 78; blue, 78 }  ,draw opacity=1 ][line width=0.75]  (206,60) -- (206,168)(242,60) -- (242,168) ; \draw  [color={rgb, 255:red, 78; green, 78; blue, 78 }  ,draw opacity=1 ][line width=0.75]  (170,96) -- (278,96)(170,132) -- (278,132) ; \draw  [color={rgb, 255:red, 78; green, 78; blue, 78 }  ,draw opacity=1 ][line width=0.75]  (170,60) -- (278,60) -- (278,168) -- (170,168) -- cycle ;
\draw [color={rgb, 255:red, 255; green, 83; blue, 13 }  ,draw opacity=1 ][line width=4.5]    (188.02,79.05) -- (218,79) -- (218,140) -- (189,140) -- (189,160) -- (230,160) -- (230,79) -- (260,79.02) ;
\draw [shift={(260,79.02)}, rotate = 0.05] [color={rgb, 255:red, 255; green, 83; blue, 13 }  ,draw opacity=1 ][fill={rgb, 255:red, 255; green, 83; blue, 13 }  ,fill opacity=1 ][line width=4.5]      (0, 0) circle [x radius= 9.05, y radius= 9.05]   ;
\draw [shift={(188.02,79.05)}, rotate = 359.91] [color={rgb, 255:red, 255; green, 83; blue, 13 }  ,draw opacity=1 ][fill={rgb, 255:red, 255; green, 83; blue, 13 }  ,fill opacity=1 ][line width=4.5]      (0, 0) circle [x radius= 9.05, y radius= 9.05]   ;
\draw  [color={rgb, 255:red, 0; green, 0; blue, 0 }  ,draw opacity=1 ][fill={rgb, 255:red, 255; green, 0; blue, 0 }  ,fill opacity=1 ] (170,96) -- (206,96) -- (206,132.05) -- (170,132.05) -- cycle ;
\draw  [draw opacity=0][fill={rgb, 255:red, 255; green, 83; blue, 13 }  ,fill opacity=1 ] (178,150) .. controls (178,144.48) and (182.48,140) .. (188,140) .. controls (193.52,140) and (198,144.48) .. (198,150) .. controls (198,155.52) and (193.52,160) .. (188,160) .. controls (182.48,160) and (178,155.52) .. (178,150) -- cycle ;

\draw (183,73) node [anchor=north west][inner sep=0.75pt]  [color={rgb, 255:red, 255; green, 255; blue, 255 }  ,opacity=1 ] [align=left] {\textbf{{\fontfamily{pcr}\selectfont 1}}};
\draw (255,73) node [anchor=north west][inner sep=0.75pt]  [color={rgb, 255:red, 255; green, 255; blue, 255 }  ,opacity=1 ] [align=left] {\textbf{{\fontfamily{pcr}\selectfont 2}}};

\end{tikzpicture}

%% file: images/experiment-transition-complexity-non-det.tex
\tikzset{every picture/.style={line width=0.75pt}} 

\begin{tikzpicture}[x=0.75pt,y=0.75pt,yscale=-1,xscale=1]

\draw  [fill={rgb, 255:red, 78; green, 78; blue, 78 }  ,fill opacity=1 ] (314,60) -- (350,60) -- (350,96.05) -- (314,96.05) -- cycle ;
\draw  [draw opacity=0][line width=0.75]  (170,60) -- (350,60) -- (350,132) -- (170,132) -- cycle ; \draw  [color={rgb, 255:red, 78; green, 78; blue, 78 }  ,draw opacity=1 ][line width=0.75]  (206,60) -- (206,132)(242,60) -- (242,132)(278,60) -- (278,132)(314,60) -- (314,132) ; \draw  [color={rgb, 255:red, 78; green, 78; blue, 78 }  ,draw opacity=1 ][line width=0.75]  (170,96) -- (350,96) ; \draw  [color={rgb, 255:red, 78; green, 78; blue, 78 }  ,draw opacity=1 ][line width=0.75]  (170,60) -- (350,60) -- (350,132) -- (170,132) -- cycle ;
\draw [color={rgb, 255:red, 18; green, 179; blue, 154 }  ,draw opacity=1 ][line width=4.5]    (188,78) -- (332,78.02) ;
\draw [shift={(332,78.02)}, rotate = 0.01] [color={rgb, 255:red, 18; green, 179; blue, 154 }  ,draw opacity=1 ][fill={rgb, 255:red, 18; green, 179; blue, 154 }  ,fill opacity=1 ][line width=4.5]      (0, 0) circle [x radius= 9.05, y radius= 9.05]   ;
\draw [shift={(188,78)}, rotate = 0.01] [color={rgb, 255:red, 18; green, 179; blue, 154 }  ,draw opacity=1 ][fill={rgb, 255:red, 18; green, 179; blue, 154 }  ,fill opacity=1 ][line width=4.5]      (0, 0) circle [x radius= 9.05, y radius= 9.05]   ;
\draw  [draw opacity=0][fill={rgb, 255:red, 18; green, 179; blue, 154 }  ,fill opacity=1 ] (286,78) .. controls (286,72.48) and (290.48,68) .. (296,68) .. controls (301.52,68) and (306,72.48) .. (306,78) .. controls (306,83.52) and (301.52,88) .. (296,88) .. controls (290.48,88) and (286,83.52) .. (286,78) -- cycle ;

\draw (183,72) node [anchor=north west][inner sep=0.75pt]  [color={rgb, 255:red, 255; green, 255; blue, 255 }  ,opacity=1 ] [align=left] {\textbf{{\fontfamily{pcr}\selectfont 1}}};
\draw (327,72) node [anchor=north west][inner sep=0.75pt]  [color={rgb, 255:red, 255; green, 255; blue, 255 }  ,opacity=1 ] [align=left] {\textbf{{\fontfamily{pcr}\selectfont 2}}};

\end{tikzpicture}

%% file: images/experiment-transition-complexity-regular.tex
\tikzset{every picture/.style={line width=0.75pt}} 

\begin{tikzpicture}[x=0.75pt,y=0.75pt,yscale=-1,xscale=1]

\draw  [fill={rgb, 255:red, 78; green, 78; blue, 78 }  ,fill opacity=1 ] (314,60) -- (350,60) -- (350,96.05) -- (314,96.05) -- cycle ;
\draw  [draw opacity=0][line width=0.75]  (170,60) -- (350,60) -- (350,132) -- (170,132) -- cycle ; \draw  [color={rgb, 255:red, 78; green, 78; blue, 78 }  ,draw opacity=1 ][line width=0.75]  (206,60) -- (206,132)(242,60) -- (242,132)(278,60) -- (278,132)(314,60) -- (314,132) ; \draw  [color={rgb, 255:red, 78; green, 78; blue, 78 }  ,draw opacity=1 ][line width=0.75]  (170,96) -- (350,96) ; \draw  [color={rgb, 255:red, 78; green, 78; blue, 78 }  ,draw opacity=1 ][line width=0.75]  (170,60) -- (350,60) -- (350,132) -- (170,132) -- cycle ;
\draw [color={rgb, 255:red, 255; green, 83; blue, 13 }  ,draw opacity=1 ][line width=4.5]    (188,78) -- (260,78) -- (260,116) -- (296,116) -- (296,78) -- (332,78.02) ;
\draw [shift={(332,78.02)}, rotate = 0.04] [color={rgb, 255:red, 255; green, 83; blue, 13 }  ,draw opacity=1 ][fill={rgb, 255:red, 255; green, 83; blue, 13 }  ,fill opacity=1 ][line width=4.5]      (0, 0) circle [x radius= 9.05, y radius= 9.05]   ;
\draw [shift={(188,78)}, rotate = 0] [color={rgb, 255:red, 255; green, 83; blue, 13 }  ,draw opacity=1 ][fill={rgb, 255:red, 255; green, 83; blue, 13 }  ,fill opacity=1 ][line width=4.5]      (0, 0) circle [x radius= 9.05, y radius= 9.05]   ;
\draw  [draw opacity=0][fill={rgb, 255:red, 255; green, 83; blue, 13 }  ,fill opacity=1 ] (286,78) .. controls (286,72.48) and (290.48,68) .. (296,68) .. controls (301.52,68) and (306,72.48) .. (306,78) .. controls (306,83.52) and (301.52,88) .. (296,88) .. controls (290.48,88) and (286,83.52) .. (286,78) -- cycle ;

\draw (183,72) node [anchor=north west][inner sep=0.75pt]  [color={rgb, 255:red, 255; green, 255; blue, 255 }  ,opacity=1 ] [align=left] {\textbf{{\fontfamily{pcr}\selectfont 1}}};
\draw (327,72) node [anchor=north west][inner sep=0.75pt]  [color={rgb, 255:red, 255; green, 255; blue, 255 }  ,opacity=1 ] [align=left] {\textbf{{\fontfamily{pcr}\selectfont 2}}};

\end{tikzpicture}

%% file: paper.bbl
\begin{thebibliography}{10}
\providecommand{\url}[1]{\texttt{#1}}
\providecommand{\urlprefix}{URL }
\providecommand{\doi}[1]{https://doi.org/#1}

\bibitem{abadi2020learning}
Abadi, E., Brafman, R.I.: Learning and solving regular decision processes. In:
  IJCAI (2020)

\bibitem{alshiekh2017safe}
Alshiekh, M., Bloem, R., Ehlers, R., Könighofer, B., Niekum, S., Topcu, U.:
  Safe reinforcement learning via shielding. In: AAAI (2018)

\bibitem{10.5555/1864519.1864559}
Bacchus, F., Boutilier, C., Grove, A.: Rewarding behaviors. In: AAAI (1996)

\bibitem{AAAI1817342}
Brafman, R., Giacomo, G.D., Patrizi, F.: {LTLf/LDLf} non-markovian rewards
  (2018)

\bibitem{brafman2019planning}
Brafman, R.I., De~Giacomo, G.: Planning for {LTLf/LDLf} goals in non-markovian
  fully observable nondeterministic domains. In: IJCAI (2019)

\bibitem{ijcai2019-766}
Brafman, R.I., De~Giacomo, G.: Regular decision processes: A model for
  non-markovian domains. In: IJCAI (2019)

\bibitem{camacho2019ltl}
Camacho, A., Icarte, R.T., Klassen, T.Q., Valenzano, R.A., McIlraith, S.A.:
  {LTL} and beyond: Formal languages for reward function specification in
  reinforcement learning. In: IJCAI (2019)

\bibitem{camacho2019learning}
Camacho, A., McIlraith, S.A.: Learning interpretable models expressed in linear
  temporal logic. In: ICAPS (2019)

\bibitem{KR2020-89}
De~Giacomo, G., Favorito, M., Iocchi, L., Patrizi, F., Ronca, A.: {Temporal
  Logic Monitoring Rewards via Transducers}. In: KR (2020)

\bibitem{RG326459658}
De~Giacomo, G., Iocchi, L., Favorito, M., Patrizi, F.: Reinforcement learning
  for ltlf/ldlf goals. arXiv preprint arXiv:1807.06333  (2018)

\bibitem{de2019foundations}
De~Giacomo, G., Iocchi, L., Favorito, M., Patrizi, F.: Foundations for
  restraining bolts: Reinforcement learning with {LTLf/LDLf} restraining
  specifications. In: ICAPS (2019)

\bibitem{10.5555/2540128.2540252}
De~Giacomo, G., Vardi, M.Y.: Linear temporal logic and linear dynamic logic on
  finite traces. In: AAAI (2013)

\bibitem{fulton2018safe}
Fulton, N., Platzer, A.: Safe reinforcement learning via formal methods: Toward
  safe control through proof and learning. In: AAAI (2018)

\bibitem{furelos2021induction}
Furelos-Blanco, D., Law, M., Jonsson, A., Broda, K., Russo, A.: Induction and
  exploitation of subgoal automata for reinforcement learning. Journal of
  Artificial Intelligence Research  \textbf{70},  1031--1116 (2021)

\bibitem{garcia2015comprehensive}
Garc{\i}a, J., Fern{\'a}ndez, F.: A comprehensive survey on safe reinforcement
  learning. Journal of Machine Learning Research  \textbf{16}(1),  1437--1480
  (2015)

\bibitem{giaquinta2018strategy}
Giaquinta, R., Hoffmann, R., Ireland, M., Miller, A., Norman, G.: Strategy
  synthesis for autonomous agents using {PRISM}. In: NASA Formal Methods
  Symposium. pp. 220--236. Springer (2018)

\bibitem{el2018personalization}
el~Hassouni, A., Hoogendoorn, M., van Otterlo, M., Barbaro, E.: Personalization
  of health interventions using cluster-based reinforcement learning. In:
  International Conference on Principles and Practice of Multi-Agent Systems
  (2018)

\bibitem{jothimurugan2020composable}
Jothimurugan, K., Alur, R., Bastani, O.: A composable specification language
  for reinforcement learning tasks. In: NeurIPS (2019)

\bibitem{kasenberg2020generating}
Kasenberg, D., Thielstrom, R., Scheutz, M.: Generating explanations for
  temporal logic planner decisions. In: ICAPS (2020)

\bibitem{kim2019bayesian}
Kim, J., Muise, C., Shah, A., Agarwal, S., Shah, J.: Bayesian inference of
  linear temporal logic specifications for contrastive explanations. In: IJCAI
  (2019)

\bibitem{nlenaers}
Lenaers, N.: An Empirical Study on Regular Decision Processes for Grid Worlds.
  Master's thesis, Department of Computer Science, Faculty of Science, Open
  University (2021)

\bibitem{li2017reinforcement}
Li, X., Vasile, C.I., Belta, C.: Reinforcement learning with temporal logic
  rewards. In: IROS (2017)

\bibitem{liao2020survey}
Liao, H.C.: A survey of reinforcement learning with temporal logic rewards
  (2020)

\bibitem{liao2020ethics}
Liao, S.M.: Ethics of artificial intelligence. Oxford University Press (2020)

\bibitem{littman2017environment}
Littman, M.L., Topcu, U., Fu, J., Isbell, C., Wen, M., MacGlashan, J.:
  Environment-independent task specifications via {GLTL}. arXiv preprint
  arXiv:1704.04341  (2017)

\bibitem{mirhoseini2021graph}
Mirhoseini, A., Goldie, A., Yazgan, M., Jiang, J.W., Songhori, E., Wang, S.,
  Lee, Y.J., Johnson, E., Pathak, O., Nazi, A., et~al.: A graph placement
  methodology for fast chip design. Nature  \textbf{594}(7862) (2021)

\bibitem{10.5555/645528.657613}
Ng, A.Y., Harada, D., Russell, S.J.: Policy invariance under reward
  transformations: Theory and application to reward shaping. In: ICML (1999)

\bibitem{van2018ethics}
van Otterlo, M.: Ethics and the value (s) of artificial intelligence. Nieuw
  Archief voor Wiskunde  \textbf{5}(19), ~3 (2018)

\bibitem{10.1109/SFCS.1977.32}
Pnueli, A.: The temporal logic of programs. In: Proceedings of the 18th Annual
  Symposium on Foundations of Computer Science (1977)

\bibitem{10.1002/9780470316887}
Puterman, M.L.: Markov Decision Processes: Discrete Stochastic Dynamic
  Programming. Wiley (1994)

\bibitem{romeo2018metric}
Romeo, {\'I}.{\'I}., Lohstroh, M., Iannopollo, A., Lee, E.A.,
  Sangiovanni-Vincentelli, A.: A metric for linear temporal logic. arXiv
  preprint arXiv:1812.03923  (2018)

\bibitem{silver2016mastering}
Silver, D., Huang, A., Maddison, C.J., Guez, A., Sifre, L., Van Den~Driessche,
  G., Schrittwieser, J., Antonoglou, I., Panneershelvam, V., Lanctot, M.,
  et~al.: Mastering the game of go with deep neural networks and tree search.
  nature  \textbf{529}(7587) (2016)

\bibitem{spaan2012partially}
Spaan, M.T.: Partially observable markov decision processes. In: Wiering, M.A.,
  {van Otterlo}, M. (eds.) Reinforcement Learning, pp. 387--414. Springer
  (2012)

\bibitem{10.5555/3312046}
Sutton, R.S., Barto, A.G.: Reinforcement Learning: An Introduction. The MIT
  Press (2018)

\bibitem{thiebaux2006decision}
Thi{\'e}baux, S., Gretton, C., Slaney, J., Price, D., Kabanza, F.:
  Decision-theoretic planning with non-markovian rewards. JAIR  \textbf{25}
  (2006)

\bibitem{van2009logic}
Van~Otterlo, M.: The Logic of Adaptive Behavior, Frontiers in Artificial
  Intelligence and Applications, vol.~192. IOS Press, Amsterdam (2009)

\bibitem{wang2019measuring}
Wang, H., Dong, S., Shao, L.: Measuring structural similarities in finite mdps.
  In: IJCAI (2019)

\bibitem{wang2020deep}
Wang, H.n., Liu, N., Zhang, Y.y., Feng, D.w., Huang, F., Li, D.s., Zhang, Y.m.:
  Deep reinforcement learning: a survey. Frontiers of Information Technology \&
  Electronic Engineering  (2020)

\bibitem{wiering2012reinforcement}
Wiering, M.A., Van~Otterlo, M.: Reinforcement learning, Adaptation, learning,
  and optimization, vol.~12. Springer (2012)

\end{thebibliography}
